\documentclass[runningheads]{llncs}

\usepackage{eccv}

\usepackage{eccvabbrv}

\usepackage{graphicx}
\usepackage{booktabs}

\usepackage[accsupp]{axessibility}  %

\usepackage{multirow}

\definecolor{Klein_Blue}{rgb}{0.0, 0.129, 0.6}
\usepackage{hyperref}       %
\hypersetup{
    colorlinks=true,
    linkcolor=magenta,
    urlcolor=magenta,
    citecolor=Klein_Blue
    }

\usepackage{orcidlink}

\newcommand{\figlabel}{Fig.\xspace}

\newcommand{\applabel}{\textbf{Appendix\xspace}}

\newcommand{\mysection}[1]{\vspace{3pt}\noindent\textbf{#1.}}

\newcommand{\minorsection}[1]{\noindent\textbf{#1.}}

\usepackage{pifont}

\definecolor{Highlight}{HTML}{39b54a}  %

\usepackage{amssymb}%
\usepackage{pifont}%

\usepackage{xcolor, colortbl}

\usepackage{algorithm}
\usepackage{listings}
\usepackage{etoolbox}
\makeatletter
\AfterEndEnvironment{algorithm}{\let\@algcomment\relax}
\AtEndEnvironment{algorithm}{\kern2pt\hrule\relax\vskip3pt\@algcomment}
\let\@algcomment\relax
\newcommand\algcomment[1]{\def\@algcomment{\footnotesize#1}}
\renewcommand\fs@ruled{\def\@fs@cfont{\bfseries}\let\@fs@capt\floatc@ruled
  \def\@fs@pre{\hrule height.8pt depth0pt \kern2pt}%
  \def\@fs@post{}%
  \def\@fs@mid{\kern2pt\hrule\kern2pt}%
  \let\@fs@iftopcapt\iftrue}
\makeatother
\newcommand{\cmmnt}[1]{}

\usepackage[normalem]{ulem}

\newcommand{\reals}{\mathbb{R}}

\newcommand{\sota}{state-of-the-art\xspace}

\usepackage{xcolor}
\definecolor{mycolor1}{HTML}{F6F6FF}
\definecolor{mycolor2}{HTML}{F6FFF6}

\newcommand{\bestr}[1]{\textcolor{BrickRed}{\textbf{#1}}}
\newcommand{\firstr}[1]{\bestr{#1}}
\newcommand{\secondr}[1]{\textcolor{NavyBlue}{\textit{#1}}}

\definecolor{deeppurple}{RGB}{255,102,255}

\usepackage{enumitem}

\newcolumntype{x}[1]{>{\centering\arraybackslash}p{#1pt}}
\newcolumntype{y}[1]{>{\raggedright\arraybackslash}p{#1pt}}
\newcolumntype{z}[1]{>{\raggedleft\arraybackslash}p{#1pt}}

\begin{document}

\title{TrackNeRF: Bundle Adjusting NeRF from Sparse and Noisy Views via Feature Tracks} 

\titlerunning{Track NeRF}

\author{
\makebox[\textwidth][c]{ Jinjie Mai\inst{1}\orcidlink{0000-0002-3396-1970} \and
Wenxuan Zhu\inst{1}\orcidlink{0009-0001-2677-3250} \and
Sara Rojas\inst{1}\orcidlink{0009-0001-4973-1694} \and
Jesus Zarzar\inst{1}\orcidlink{0000-0003-4966-9621}  \and}
\makebox[\textwidth][c]{ Abdullah Hamdi\inst{2}\orcidlink{0000-0003-3989-7540} \and
Guocheng Qian\inst{3}\orcidlink{0000-0002-2935-8570} \and
Bing Li\inst{1}\orcidlink{0000-0001-9465-8142} \and}
\makebox[\textwidth][c]{Silvio Giancola\inst{1}\orcidlink{0000-0002-3937-9834} \and
Bernard Ghanem\inst{1}\orcidlink{0000-0002-5534-587X}}
}

\authorrunning{J.~Mai et al.}

\institute{King Abdullah University of Science and Technology \and Visual Geometry Group, University of Oxford \and Snap Inc. \\
\email{\{jinjie.mai,bernard.ghanem\}@kaust.edu.sa}
}

\makeatletter
\let\@oldmaketitle\@maketitle%
\renewcommand{\@maketitle}{\@oldmaketitle%
\myfigure\bigskip}
\makeatother
\newcommand\myfigure{%
  \makebox[0pt]{\hspace{\textwidth}\includegraphics[width=1.0\textwidth]{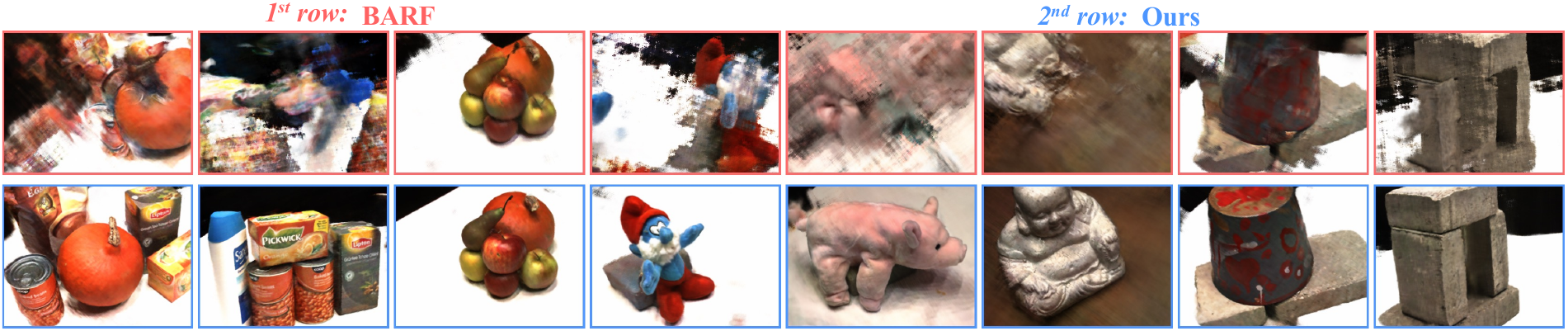}}
  \\
\refstepcounter{figure}\textbf{Fig.~\thefigure}: {\small \textbf{Novel View Synthesis from Sparse and Noisy Views.} TrackNeRF achieves high-quality novel view synthesis through \textit{bundle adjusting feature tracks. } } 
  \label{fig:teaser}
}

\maketitle

\begin{abstract}
Neural radiance fields (NeRFs) generally require many images with accurate poses for accurate novel view synthesis, which does not reflect realistic setups where views can be sparse and poses can be noisy.
Previous solutions for learning NeRFs with sparse views and noisy poses only consider local geometry consistency with pairs of views. 
Closely following \textit{bundle adjustment} in Structure-from-Motion (SfM), we introduce TrackNeRF for more globally consistent geometry reconstruction and more accurate pose optimization. TrackNeRF introduces \textit{feature tracks}, \ie connected pixel trajectories across \textit{all} visible views that correspond to the \textit{same} 3D points. By enforcing reprojection consistency among feature tracks, TrackNeRF encourages holistic 3D consistency explicitly. Through extensive experiments, TrackNeRF sets a new benchmark in noisy and sparse view reconstruction.
In particular, TrackNeRF shows significant improvements over the state-of-the-art BARF and SPARF by $\sim8$ and $\sim1$ in terms of PSNR on DTU under various sparse and noisy view setups.
The code is available at \href{https://tracknerf.github.io/}{\texttt{\textcolor{purple}{\uline{https://tracknerf.github.io/}}}}.

\end{abstract}

\section{Introduction}
\label{sec:intro}

The pursuit of reconstructing and creating immersive virtual environments, such as the metaverse, has accelerated significantly with the advent of 3D vision and AR/VR devices. 
Creating such realistic virtual environments typically requires intricate manual design by artists.
Concurrently, the evolution of novel-view synthesis has been notably influenced by the emergence of Neural Radiance Fields (NeRFs)~\cite{NeRF}.
NeRFs have revolutionized this domain by parameterizing the volume of scenes through neural networks.
They enable rendering novel views of a scene with unparalleled precision and realism, thus emulating a virtual environment.
This foundational shift towards leveraging neural networks lays the groundwork for the next generation of immersive digital experiences, driving forward the boundaries of what is possible in virtual world creation.

\begin{figure}[t]
    \centering
    \includegraphics[width=\linewidth]{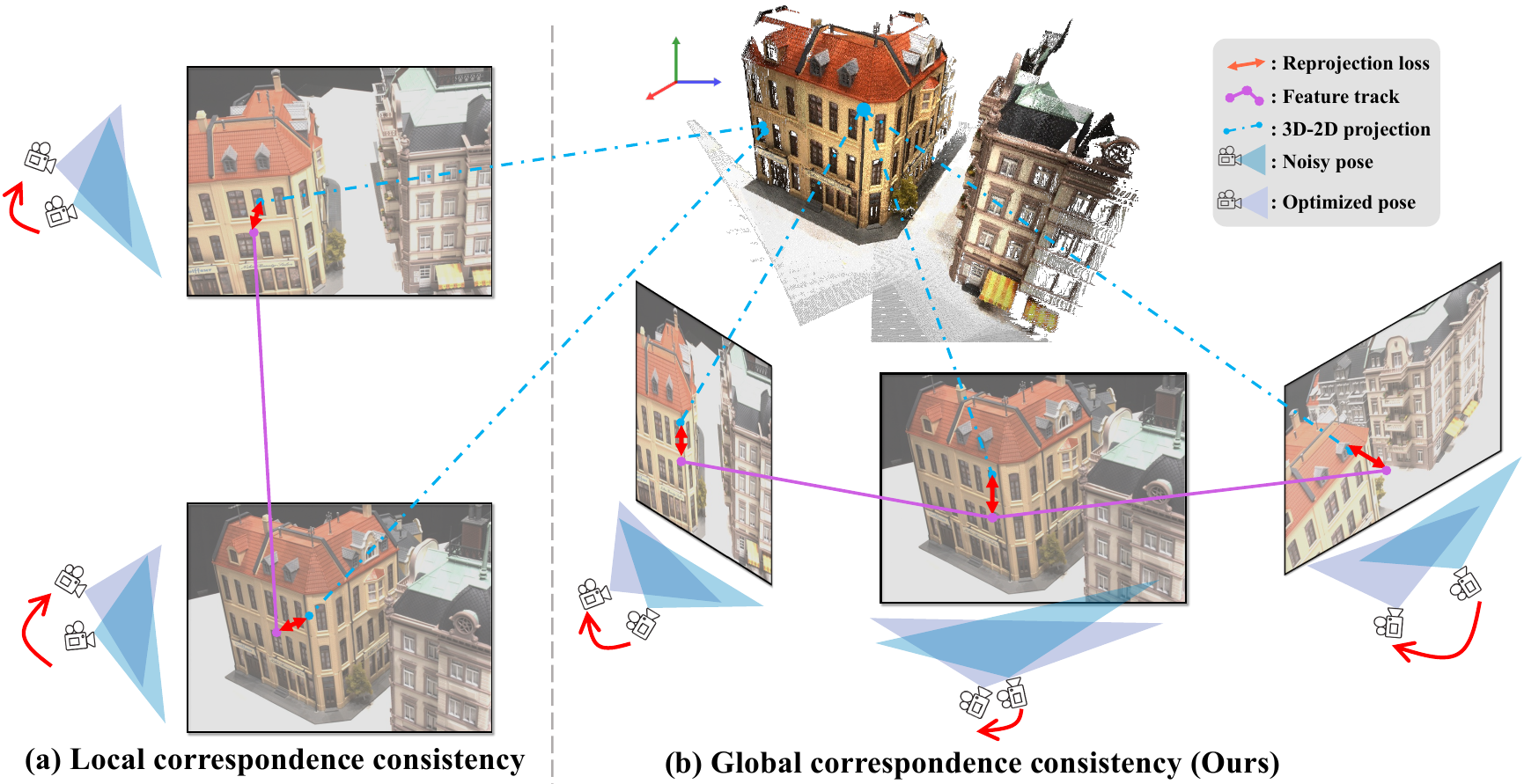}
    \caption{
    \textbf{Illustration of Track Reprojection Loss.} \textit{Left:} Pairwise correspondence objective employed by CorresNeRF~\cite{corresnerf} and SPARF~\cite{sparf2023}. \textit{Right:} Feature tracks objective proposed by \textbf{TrackNeRF}. 
    TrackNeRF minimizes the reprojection loss across all visible views for feature tracks corresponding to the same landmarks.
    }
    \label{fig:pipeline}
\end{figure}

Neural Radiance Fields and their derivatives, including prominent methods like InstantNGP~\cite{InstantNGP} and Plenoxels~\cite{plenoxels}, along with recent advancements in Gaussian Splatting~\cite{gaussiansplatter}, fundamentally rely on the availability of a substantial corpus of posed multiview images to attain photorealistic renderings.
This assumption, however, starkly contrasts with the typical conditions encountered in real-world scenarios. 
In most cases, available visual data is comprised of sparse image collections or unposed videos with standard Structure-from-Motion (SfM) tools such as COLMAP~\cite{colmap_sfm,egoloc} employed to derive camera poses.
Both scenarios inevitably result in estimations tarnished by noise and inaccuracies. 
This discrepancy between the ideal conditions presumed by NeRF-based methodologies and the reality of data acquisition presents a significant challenge. 
Novel approaches that efficiently work with less-than-ideal datasets are necessary to ensure robust 3D reconstruction and rendering without the luxury of extensive posed imagery.

Previous works~\cite{PixelNeRF,dsnerf,nerfminusminus,camp,reconfusion,freenerf,corresnerf,consistent_nerf} have attempted to address this more realistic setup.
BARF~\cite{barf} first proposes adopting coarse-to-fine frequency encoding to ease NeRF's backward optimization for pose estimation. %
However, BARF does not perform de facto \textit{``bundle adjustment''} since it does not use any multiview constraints.
Other methods tackle noisy poses as well by assuming dense views~\cite{nerfminusminus,barf,camp} or address the few-view limitation by assuming perfect poses~\cite{PixelNeRF,reconfusion}, but don't tackle both issues jointly. 
Recently, SPARF~\cite{sparf2023} proposes to tackle sparse views and noisy poses simultaneously, achieving a remarkable milestone in the development of realistic setup novel views synthesis. 
However, SPARF adopts two-view correspondence-based reprojection loss as the core of optimization without considering long-term consistency.

To overcome the local consistency limitation of previous works, we propose TrackNeRF. TrackNeRF is inspired by the fact that all views are taken from \textit{a single holistic 3D scene}, thus the corresponding pixels from \textit{all} sparse views rendered from NeRF should ideally be back-projected to the \textit{same} 3D landmark. Following this motivation, TrackNeRF extracts \textit{feature tracks} for optimization, \ie connected pixel trajectories across all visible views corresponding to the same 3D points. 
TrackNeRF enforces reprojection consistency among each track and thus introduces a holistic geometry consistency into NeRF. 
As a result, TrackNeRF achieves the most robust and highest reconstruction fidelity with more precise camera poses among all sparse and noisy NeRF solutions.
\figlabel\ref{fig:teaser} illustrates the difference between TrackNeRF and the pioneering methods.

\noindent\textbf{Contributions.} We summarize our contributions as follows:

\noindent\textbf{(i)} We introduce TrackNeRF, which utilizes feature tracks that closely follow the \textit{bundle adjustment} literature. 
TrackNeRF can reconstruct a more geometry-consistent volumetric representation and recover more accurate camera poses.

\noindent\textbf{(ii)} Our TrackNeRF achieves $\sim1$ PSNR boost against previous SOTA and halves pose errors on the challenging DTU~\cite{dtu} dataset for all 3-view, 6-view, and 9-view setups with noisy poses.
Under 3 views with ground truth poses, TrackNeRF also outperforms SOTA diffusion and regularization-based methods.

\noindent\textbf{(iii)} In practice, we demonstrate that TrackNeRF can tolerate greater pose noise, perform faster pose optimization, and synthesize high-quality novel views aligned with correct and smooth depth.

\section{Related Work}

\minorsection{Multi-View Reconstruction}
The multi-view 3D reconstruction field is dedicated to the restoration of a scene's three-dimensional structure from its two-dimensional RGB images, obtained from various camera perspectives~\cite{UncalibratedStereoRig, RomeInADay}. Historically, these approaches have focused on generating a point cloud representation of a scene's geometry through the utilization of SIFT-based point matching techniques~\cite{SIFT, colmap_sfm, colmap_mvs, egoloc}. Advances in this domain have seen a shift towards the use of neural networks to improve feature extraction, as evidenced by several studies (\eg \cite{mvsnet, DeepMVS, rmvsnet, fastmvsnet,voint,ET-MVSNet}). The introduction of Neural Radiance Fields (NeRF)~\cite{NeRF, NeuralVolumes} has marked a significant transition toward the volumetric radiance reconstruction of 3D spaces~\cite{VolumeRenderingDigest}, facilitating the creation of photorealistic novel views~\cite{RefNeRF, MipNeRF, MipNeRF-360}. However, as a common constraint, all of these approaches assume \textit{sufficient overlapped views}, typically around $100$ images, with \textit{precise camera poses}. In many real applications, the captured views are sparse and noisy, consisting of only two to four views with inaccurate poses. Our aim in this work is to robustly reconstruct 3D geometry from these sparse and noisy poses. 

\minorsection{Few-View Reconstruction}
Further research has delved into optimizing NeRF (\eg \cite{DietNeRF, InfoNeRF, WideBaseline,Regnerf,sparsenerf,sparf_abd,match_nerf,consistent_nerf,cmc_nerf,re_nerfing}) and Gaussian Splatting (\eg~\cite{gaussiansplatter,ges,li2024dngaussian,chen2024mvsplat,xiong2023sparsegs,zhu2023fsgs}) for scenarios with limited shots and even single-shot contexts (\eg PixelNeRF \cite{PixelNeRF}), focusing on density fields without explicit 3D geometrical storage. 
DS-NeRF \cite{dsnerf} regulates the NeRF rendering with monocular depth estimation, enhancing the quality and speed of few-view optimization. 
SfMNeRF~\cite{chen2023improving} optimizes the left-right reprojection loss in 3D and also applies the depth smoothness term.
FreeNeRF \cite{freenerf} regularizes the input frequency of the NeRFs and the occluded regions to improve the performance at the few-views setup.   
The evolution of zero-shot single-view 3D reconstruction has been significantly propelled by advances in multi-modal diffusion models and zero-shot 3D synthesis technologies~\cite{DreamFusion, Magic3D, Fantasia3D, DITTO-NeRF, RealFusion, Zero-1-to-3, Magic123,charatan2024pixelsplat,szymanowicz2024splatter}. 
Recently, ReconFusion \cite{reconfusion} and Zero-MVS \cite{zeronvs} utilized the 2D diffusion prior to greatly enhance the quality of novel view synthesis with very few views, achieving \sota in this domain. Our TrackNeRF does not use any generative priors to enhance the sparse view reconstruction but rather relies on generic geometric cues and feature tracks from the same scene that are more generalizable and tackle both sparse \textit{and} noisy poses.

\minorsection{NeRFs with no Pose Requirements}
Many works~\cite{niceslam,nerfslam,Regnerf,camp,SCNeRF,MipNeRF-360,bad_nerf,l2g_ba_nerf,usb_nerf,uc_nerf,GARF,cbarf,dbarf,coponerf,up_nerf,chen2022structure,sun2023icomma,fu2023colmapfree} have tried to optimize NeRF from noisy pose or without pose.
NeRF$--$~\cite{nerfminusminus} is a pioneering work to investigate fitting a NeRF jointly with camera poses for novel view synthesis.
BARF \cite{barf} adds frequency modulation coordinate embedding in NeRF to greatly enhance camera optimization in noisy pose cases.
SPARF \cite{sparf2023} introduces a pairwise correspondence loss to the NeRF formulation and shows great performance in the case of \textit{jointly} noisy and sparse views, a combined setup that was not properly performed previously. Our TrackNeRF is different because it assumes a global geometry loss that constrains \textit{all} the cameras jointly based on the feature tracks.

\section{Preliminaries}

\minorsection{NeRF}
Given a set of sparse views and associated camera parameters,  NeRF~\cite{NeRF} learns an implicit neural representation to represent the 3D scene.
Let $R_i \in SO(3)$ and $\mathbf{t}_i\in \reals^{3}$ denote the rotation and translation of camera pose $i$, respectively. 
The camera-to-world transformation  of  the $i$-th camera is denoted   as $P_i = \left[R_i|\mathbf{t}_i\right] \in SE(3)$.

To render a pixel $\mathbf{p} \in \reals^{2}$ from a given camera with pose $P_{i}$, NeRF traces a ray from the projection of the camera center $\mathbf{t}_i$ along the direction defined by $\mathbf{d}_{i, \text{p}} = R_i K_i^{-1}\bar{\mathbf{p}}$  in the world coordinate system where $K \in \mathbb{R}^{3\times3}$ is the camera's intrinsic matrix and $\bar{\mathbf{p}}$ is homogeneous representation of ${\mathbf{p}}$.
We then discretely sample $M$  points along the ray, bounded by the near and far planes, to  predict the color $\hat{\mathbf{I}}_{i, \text{p}}$ of a pixel from the radiance field as:
\begin{align}
\vspace{-1mm}
\hat{\mathbf{I}}_{i, \text{p}} &= \hat{I}(\mathbf{p}; \theta, P_i) = \sum_{m=1}^{M} \alpha_m\mathbf{c}_m \,, \label{eq:volume_rendering} 
\end{align}
\text{where}  $\hat{I}$ is the  RGB  rendering function, $\left\{(\mathbf{c}_m, \sigma_m  ) \right\}_{m=1}^M$ are the color and volume density of sampled points predicted by a radiance field parameterized by $\theta$.
Let $\hat{z}(\cdot)$ be the depth rendering functions and $z_m$ be the ray depth at sampled point $m$. We approximately estimate the depth of the scene perceived from  $\mathbf{p}$ as,
\begin{equation}
\vspace{-1mm}
\label{eq:rendered-depth}
    \hat{z}_{i, \text{p}} = \hat{z}(\mathbf{p}; \theta, P_i) =  \sum_{m=1}^{M} \alpha_m z_m \,. 
\vspace{-1mm}
\end{equation}

\minorsection{Photometric loss} 
NeRF approaches~\cite{barf, nerfminusminus, GARF,sinerf,sparf2023,corresnerf} typically use a photometric loss to optimize radiance field parameters $\theta$ as well as camera poses $\hat{\mathcal{P}}$.
Let $\hat{P}_i$ denote the pose estimate for the $i$-th training image.
The photometric loss is defined as follows:
\begin{equation}
\label{eq:photo-pose-theta}
    \mathcal{L}_{\text{Photometric}}(\theta, \hat{\mathcal{P}}) = \frac{1}{n} \sum_{i=1}^n \sum_{\text{p}} \left\| I_i(\mathbf{p}) - \hat{I}(\mathbf{p}; \theta, \hat{P}_i) \right\|_2^2 \,.
\end{equation}
where $n$ is the number of training images.
Different from these approaches, we propose a track reprojection loss that effectively reduces the negative effects of noisy camera poses on NeRF results.

\section{TrackNeRF} \label{sec:method}

Given the fact that all sparse views are shot from a single holistic 3D scene, corresponding points from \textit{all} views rendered from the 3D model should ideally be projected back to the same 3D landmark.
Unfortunately, even the most recent state-of-the-art methods SPARF \cite{sparf2023} and CorresNeRF \cite{corresnerf,consistent_nerf} only consider local matching consistency from a pair of renderings to train NeRF from sparse and noisy views.
These works fail to exploit holistic consistency across all views.
Inspired by the bundle adjustment from Structure-from-Motion~\cite{colmap_sfm,pixel-perfect-sfm}, our work follows the track-wise objective of BA instead:
\begin{align}
E_{BA}=\sum_k \sum_{(\mathbf{u}_i,\mathbf{v}_j)\in \mathcal{T}_{(k)}} \| h (\mathbf{u}_i) - \mathbf{v}_j \| \label{eq:sfm_ba}
\end{align}
where $(\mathbf{u}_i,\mathbf{v}_j)$ is a correspondence inside feature track $\mathcal{T}_{k}$ between pixels $\mathbf{u}_i$ and $\mathbf{v}_j$ from views $i$ and $j$ respectively.
The function $h$ lifts a pixel onto its 3D location and projects it to a different view. 
Note that $\mathcal{T}_{k}$ considers correspondences between points across all visible views rather than pairs of views as in~\cite{sparf2023,corresnerf}.  
This track-wise loss encourages all pixels in a feature track to correspond to the same 3D landmark and enforces holistic consistency. 
In our paper, we propose to jointly optimize the matching correspondence for the whole feature track that aligns with the concept of \textit{bundle adjustment} (BA) in Eq.~\ref{eq:sfm_ba}. 
This differs our work from BARF~\cite{barf}, which only considers position encoding strategy without multiview correspondences and objective of BA.
We elaborate on the formalization of this track-wise objective in the context of NeRF and details of our method next.

\subsection{Track Adjustment}
\mysection{Track Extraction} Compared to \cite{karaev2023cotracker,doersch2023tapir,vggsfm}, we extract tracks by chaining. Given a set of images $\{\mathcal{I}\}$ of size $N$ and a feature matcher~\cite{pdcnet++} $\mathcal{F}$, we can extract feature correspondences for every image pair.
If a correspondence $(\mathbf{u},\mathbf{v})$ between $\mathcal{I}_i$ and $\mathcal{I}_j$ is found,
and there is also a correspondence $(\mathbf{v},\mathbf{q})$ between $\mathcal{I}_j$ and $\mathcal{I}_k$,
then it implies a transitive relationship that can be extended to form a continuous feature track.
By connecting all such correspondences, we can get feature tracks $\{\mathcal{T}\}$, with $\mathcal{T}_k = \{\mathbf{u}, \mathbf{v}, \mathbf{q}, ...\} $, where $\mathbf{u},\mathbf{v},\mathbf{q}$ are pixel coordinates from different images corresponding to the same 3D point.
Note that although we are using a dense correspondence model~\cite{pdcnet++}, we adopt the term \textit{``feature"} in our paper to align with the concept of \textit{``track"}.

\mysection{Track Keypoint Adjustment} Recent advances~\cite{pixel-perfect-sfm,particlesfm} in SfM~\cite{colmap_sfm} have shown significant benefits by optimizing feature tracks before triangulation and bundle adjustment. 
To obtain more accurate feature tracks,  for each track $\mathcal{T}_k$, we adhere to PixSfM~\cite{pixel-perfect-sfm} by doing track-wise keypoint adjustment to encourage multiview consistency:

\begin{align}
    E^k_{TKA}=\sum_{(\mathbf{u}_i,\mathbf{v}_i)\in \mathcal{T}_{(k)}} w_{\mathbf{u}_i\mathbf{v}_j} \| \mathbf{F}(\mathbf{u}_i) - \mathbf{F}(\mathbf{v}_j) \| \label{eq:tka}
\end{align}
where $\mathbf{F}$ is the function that extracts pixel-wise features for keypoints, and $w_{\mathbf{u}_i\mathbf{v}_j} $ is the matching confidence from our matcher ~\cite{pdcnet++}.
The objective of Eq.~\ref{eq:tka} will refine the location of matched keypoints by minimizing the difference of pixel-wise features through traceable numeric gradients.
Since the refinement optimizes the feature-metric consistency inside the whole feature track, we obtain more accurate keypoints for the supervision of NeRF.

\subsection{Track Reprojection Loss}
Previous methods~\cite{sparf2023, corresnerf,consistent_nerf,chen2023improving} only exploit optimizing NeRF with pairwise correspondence from two views independently during each iteration.
However, feature extractors~\cite{SIFT,superpoint,pdcnet++} working on single view are easily perturbed by appearance variations~\cite{benchmarking_vloc} thus may introduce multiview inconsistency for matching and reconstruction~\cite{mv_optimization_local_geometry}.
To alleviate this, we propose our track reprojection loss to enforce global geometric consistency for the radiance field,
where multiview geometry constraints for the feature tracks are optimized simultaneously.

The track projection loss is based on the principle of bundle adjustment in Eq.~\ref{eq:sfm_ba}, \ie, all correspondences in a feature track $\mathcal{T}_k$ should correspond to the same 3D landmark (see Fig. \ref{fig:pipeline}). In particular,  
let  $(\mathbf{u}_i,\mathbf{v}_j)$ be a pair of matching pixels between image $\mathcal{I}_i$ and $\mathcal{I}_j$ sampled from a feature track $\mathcal{T}_k$, where $\mathbf{v}_j$ is a pixel in  $\mathcal{I}_j$.  For $\mathbf{v}_i$ and its depth estimated by Eq. \ref{eq:rendered-depth}, we can obtain its corresponding 3D point in the world coordinate system using camera projection and camera-to-world transformation. We then reproject the corresponding 3D point of $\mathbf{v}_j$ to the image plane of image $\mathcal{I}_i$.
Since $(\mathbf{u}_i,\mathbf{v}_j)$ corresponds to the same 3D point in the world coordinate system, the reprojection of $\mathbf{v}_j$ should overlap with pixel $\mathbf{u}_i$. 
Therefore, given the sampled feature track $\mathcal{T}_k$ optimized by Eq.~\ref{eq:tka} and consisting of \textit{all-to-all} correspondences like $(\mathbf{u}_i,\mathbf{v}_j)$, our track reprojection loss minimizes the overall distance between all such $\mathbf{u}_i$ and the reprojection of $\mathbf{v}_j$ inside $\mathcal{T}_k$ as follows:
\begin{align}
\mathcal{L}_{\text{Track}} = \sum_k \sum_{(\mathbf{u}_i,\mathbf{v}_j)\in \mathcal{T}_k} \frac{1}{| \mathcal{T}_k| } \rho \left( \, \mathbf{u}_i - \pi \left(  \hat{P}_i^{-1} \hat{P}_j  \,\,\,  \pi^{-1}   (  \mathbf{v}_j, \hat{z}(\mathbf{v}_j; \theta, \hat{P}_i) \, )   \right) \right) \,
\label{eq:track_loss}
\end{align}
where $\rho$ is the Huber loss function~\cite{hastie01statisticallearning}, $\pi$ is the camera projection operator which maps a 3D point in the camera coordinate system to the image plane, $\pi^{-1}$ is the backprojection operator which projects a pixel $v_j$ back to the camera coordinate system using the pixel's depth $\hat{z}(\mathbf{v}_j; \theta, \hat{P}_i)$ estimated by Eq. \ref{eq:rendered-depth}, and $\hat{P}_i^{-1}$ projects a 3D point in the world coordinate system to the camera coordinate system of image $\mathcal{I}_i$.

\subsection{Depth Regularization}

NeRF~\cite{NeRF} is trained only using photometric loss, so it suffers from poor geometry, especially in sparse view settings.
As reported by previous methods~\cite{floater_no_more,Regnerf,freenerf}, depth ambiguity results in many floaters in the trained NeRF and significantly degrades the quality of novel view synthesis.
To regularize the internal geometry of the radiance field, 
we also introduce a depth regularization loss to encourage depth gradients to align with rendered image gradients following the practice of monocular depth estimation~\cite{pmhuber,geonet,godard2017unsupervised}:

\begin{align}
    \mathcal{L}_{\text{Depth}} = \sum_{i,j}^{\Psi} | \nabla_x D^{\Psi}_{ij} | e^{-\| \nabla_x I^{\Psi}_{ij} \| } + | \nabla_y D^{\Psi}_{ij} | e^{-\| \nabla_y I^{\Psi}_{ij} \| }
    \label{eq:depth_reg}
\end{align}

where $\nabla$ is the vector differential operator, $D$ is the disparity map by taking the reciprocal of Eq.~\ref{eq:rendered-depth}, $I$ is the RGB image, and $\Psi$ represents the sampled image patch for regularization. 

\subsection{NeRF training}

Combining the L2 photometric loss, the depth regularization loss, and the proposed track reprojection loss, we optimize our TrackNeRF from sparse views with possibly noisy poses as follows:
\begin{align}
    \mathcal{L} = \mathcal{L}_{\text{Photometric}} + \lambda_{\text{Depth}} \mathcal{L}_{\text{Depth}} + \lambda_{\text{Track}} \mathcal{L}_{\text{Track}}
    \label{eq:loss_all}
\end{align}
where $\lambda_{Depth}$ and $\lambda_{Track}$ are weighting factors of $\mathcal{L}_{\text{Depth}}$ and $\mathcal{L}_{\text{Track}}$, respectively.

\begin{figure}[tb]
    \centering
    \includegraphics[width=1.\textwidth]{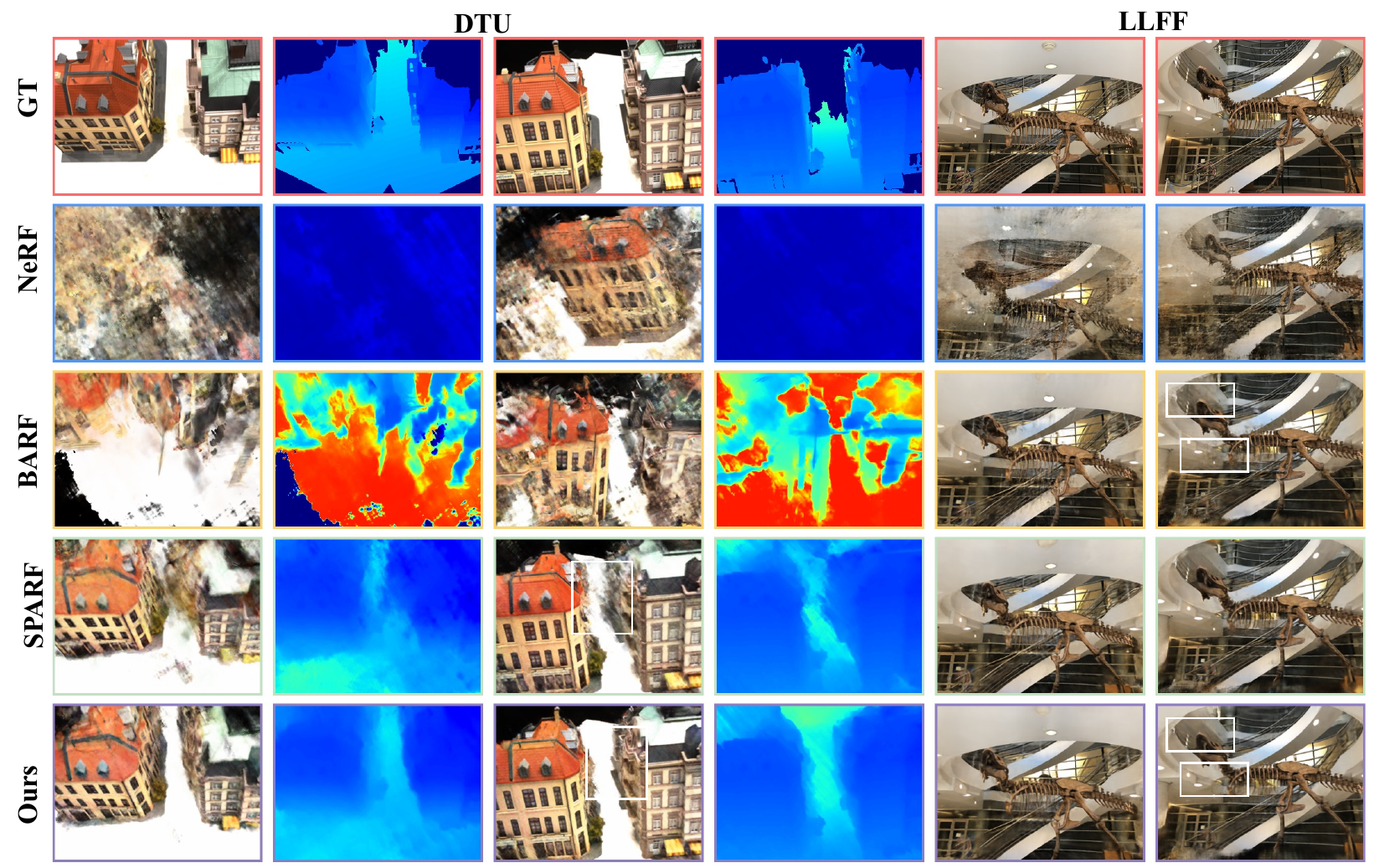}
    \caption{
    \textbf{Quanlitative Comparison on DTU~\cite{dtu} and LLFF~\cite{llff}.} We show views from the test view split of both datasets to visually compare our TrackNeRF renderings to the baselines. For DTU dataset where GT depth maps are available, we additionally visualize the rendered depth by Eq.~\ref{eq:rendered-depth} to compare the learned geometry.
    }

    \vspace{-6pt}
    \label{fig:vis_main}
\end{figure}

\section{Experiments and Results}\label{sec:exp}

\subsection{Experiment Setups}
\noindent\textbf{Datasets and Metrics.}
We extensively evaluate our proposed TrackNeRF on \textbf{DTU}~\cite{dtu} and \textbf{LLFF}~\cite{llff} datasets under various settings. 
DTU is a challenging benchmark as there are usually wide baselines between different views.
We follow the split of PixelNeRF~\cite{PixelNeRF} to conduct our evaluation on the test split of 15 scenes,
while we also report the results with masking background as done by~\cite{sparf2023,freenerf,reconfusion}.
For the LLFF dataset, we select every $8^{th}$ image for testing as NeRF~\cite{NeRF}.
Regarding the metrics, we adopt PSNR, SSIM~\cite{ssim}, and LPIPS~\cite{LPIPS,alexnet} following the community standard.
Since ground truth depth maps are available on DTU, we also report the mean depth absolute error (DE) like SPARF~\cite{sparf2023}.

\noindent\textbf{Implementation details.}
We follow the training strategy of SPARF~\cite{sparf2023} for a fair comparison, where we jointly optimize NeRF with poses at the first stage and then only finetune NeRF at the second stage.
We also utilize the same correspondence network, PDCNet++~\cite{pdcnet++}, as SPARF for fairness.
We adopt 6-DoF camera pose representation in ~\cite{continuity_pose}.
We sample random rays for photometric loss, random feature tracks for track loss, and random pixel patches for depth loss.
The depth regularization loss is only enabled in the second stage after pose optimization.

\begin{table}[tb]
\centering
\caption{\textbf{DTU Evaluation (3, 6 \& 9 Noisy Views)}. We evaluate methods for unseen view rendering and camera extrinsic recovery on the DTU dataset \cite{dtu}, using initial poses that are noisy and vary in terms of the number of input views (3, 6, or 9). We introduce noise to these poses by adding $15\%$ Gaussian noise to the true poses. The rotation errors are measured in degrees, while the translation errors are scaled by a factor of 100. The results in parentheses ($\cdot$) are obtained after applying a mask to the background.
Our TrackNeRF achieves the best performance in all setups. 
}
\resizebox{1.0\textwidth}{!}{%
\begin{tabular}{l|l|cc|cccc}
\toprule
 & \textbf{Method} & \textbf{Rot. $\downarrow$} & \textbf{Trans. $\downarrow $} & \textbf{PSNR $\uparrow$} & \textbf{SSIM $\uparrow$} & \textbf{LPIPS $\downarrow$} & \textbf{DE $\downarrow $} \\
 \midrule
\multirow{6}{*}{\rotatebox[origin=c]{90}{\textbf{3 input views}}}
 & BARF~\cite{barf}                    & 10.33 & 51.5 & 10.71 (9.76) & 0.43 (0.62) & 0.59 (0.36) & 1.90 \\ 
 & RegBARF~\cite{barf, Regnerf}        & 11.20 & 52.8 & 10.38 (9.20) & 0.45 (0.62) & 0.61 (0.38) & 2.33\\
 & DistBARF~\cite{barf, MipNeRF-360}   & 11.69 & 55.7 & 9.50 (9.15) & 0.34 (0.76) & 0.67 (0.36) & 1.90\\
 & SCNeRF~\cite{SCNeRF}                & 3.44 & 16.4 & 12.04 (11.71) & 0.45 (0.66) & 0.52 (0.30) & 0.85 \\
 & SPARF~\cite{sparf2023}              & \secondr{1.81} & \secondr{5.0}& \secondr{17.74 (18.92)} & \secondr{0.71 (0.83)} & \secondr{0.26 (0.13)} & \secondr{0.12} \\
 & \textbf{TrackNeRF (Ours)}           & \firstr{1.12} & \firstr{2.48} & \firstr{18.53 (19.65}) & \firstr{0.73 (0.83)} & \firstr{0.25 (0.13)} & \firstr{0.11} \\
 \midrule
\multirow{6}{*}{\rotatebox[origin=c]{90}{\textbf{6 input views}}}
 & BARF~\cite{barf}                    & 9.20 &31.1 & 14.02 (14.22) & 0.54 (0.69) & 0.46 (0.27 ) & 0.49  \\
 & RegBARF~\cite{barf, Regnerf}        & 9.19 &26.63 & 14.59 (14.58) & 0.57 (0.70) & 0.44 (0.27) & 0.32 \\
 & DistBARF~\cite{barf, MipNeRF-360}   & 8.96 &28.85 & 14.31 (14.60) & 0.55 (0.70) & 0.43 (0.26) & 0.53 \\
 & SCNeRF~\cite{SCNeRF}                & 4.10 &12.80 & 17.76 (18.16) & 0.70 (0.80) & 0.31 (0.18) & 0.28  \\
 & SPARF~\cite{sparf2023}              & \secondr{1.31} &\secondr{2.7} & \secondr{21.39} (\secondr{22.01}) & \secondr{0.81} (\secondr{0.88}) & \secondr{0.18} (\secondr{0.10}) & \secondr{0.09} \\
 & \textbf{TrackNeRF (Ours)}           & \firstr{0.24} &\firstr{0.65} & \firstr{22.78 (23.66)} & \firstr{0.84 (0.89)} & \firstr{0.14 (0.08)} & \firstr{0.06} \\
 \midrule
\multirow{6}{*}{\rotatebox[origin=c]{90}{\textbf{9 input views}}}
 & BARF~\cite{barf}                    &  8.34 &26.72 & 16.20 (16.38) & 0.60 (0.73) & 0.38 (0.22) & 0.35 \\
 & RegBARF~\cite{barf, Regnerf}        &5.28 &18.51 & 18.98 (19.08) & 0.67 (0.77) & 0.29 (0.18) & 0.23 \\
 & DistBARF~\cite{barf, MipNeRF-360}   &7.00 &26.42 & 16.18 (16.27) & 0.58 (0.71) & 0.37 (0.22) & 0.29 \\
 & SCNeRF~\cite{SCNeRF}                &  4.76 &16.25 & 18.19 (18.01) & 0.69 (0.81) & 0.31 (0.17) & 0.31 \\
 & SPARF~\cite{sparf2023}              &  \secondr{1.15} &\secondr{2.55} & \secondr{24.69} (\secondr{25.05}) & \secondr{0.88} (\secondr{0.92}) & \secondr{0.12}(\secondr{0.06}) & \secondr{0.06} \\ 
 & \textbf{TrackNeRF (Ours)}           &\firstr{0.25} &\firstr{0.70} & \firstr{25.57 (26.03)} & \firstr{0.89 (0.92)} & \firstr{0.11 (0.06)} & \firstr{0.05} \\ 
\bottomrule
\end{tabular}%
}

\label{tab:dtu_noise_pose}
\end{table}

\subsection{Main Results}
\minorsection{DTU noisy 3 views} We keep the pose setting from previous methods~\cite{barf,sparf2023} by adding $15\%$ additive gaussian noise to the groundtruth camera poses.
As shown in Tab.~\ref{tab:dtu_noise_pose}, TrackNeRF achieves \sota performance across all the metrics.
Specifically, for pose optimization, TrackNeRF nearly halves both the rotation and translation errors. 
It also improves PSNR by a remarkable margin of $\sim 0.8$.

\minorsection{DTU noisy 6 views and 9 views} We extend similar settings to 6-view and 9-view cases, the results of which are presented in Tab.~\ref{tab:dtu_noise_pose}.
Surprisingly, the biggest improvement is observed when there are \textit{moderate sparse} views, where a $1.65$ boost on PSNR and huge reduction of pose error can be observed in the 6-view case.
We believe such improvements benefit from multiview consistency from longer feature tracks.
When the views become denser, we can still achieve $1$ PSNR boost and smaller pose drift in the 9-view case.

\begin{table}[tb]
\centering
\caption{\textbf{DTU Evaluation with Ground Truth Poses (3 Views)}. We show the evaluation on DTU \cite{dtu} with three input views and \textit{ground truth poses}, where the first column lists the schemes of methods and their architectures (\textbf{VN}: Vanilla NeRF , \textbf{MN}: MipNeRF, \textbf{PN}: PixelNeRF , \textbf{D}: Diffusion). } 
{%
    \tabcolsep=0.08cm
\begin{tabular}{ll|c|ccc}
\toprule

\textbf{\multirow{2}{*}{Archit.}} & \textbf{\multirow{2}{*}{Method}} & \textbf{\multirow{ 2}{*}{Settings}} & 
\textbf{PSNR $\uparrow$} & \textbf{SSIM $\uparrow$ } & \textbf{LPIPS  $\downarrow$} \\
& & & \textbf{(masked) } & \textbf{ (masked)} & \textbf{  (masked)} \\
\midrule
PN   & PixelNeRF \cite{PixelNeRF} & Trained        & \firstr{19.36} (18.00) & 0.70 (0.77) & 0.32 (0.23) \\
PN+D & ReconFusion \cite{reconfusion} & on  & - (20.74) & - (0.88) & - (0.12) \\
D   & ZeroNVS \cite{zeronvs}   & datasets       & - (16.71) & - (0.72) & - (0.22) \\ \midrule
MN     & MipNeRF \cite{MipNeRF}  &               & - (16.11) & - (0.40) & - (0.46) \\
MN     & RegNeRF \cite{Regnerf}   &               & - (18.84) & - (0.57) & - (0.36) \\
MN     & FreeNeRF \cite{freenerf}  & Optimized     & - (20.46) & - (0.83) & - (0.17) \\
VN        & DS-NeRF \cite{dsnerf}    & per           & 16.52 (-) & 0.54 (-) & 0.48 (-) \\
VN        & CorresNeRF \cite{corresnerf} & scene         & 18.23 (20.58) & 0.76 (0.77) & 0.33 (0.13) \\
VN        & SPARF \cite{sparf2023}     &               & 18.30 \secondr{(21.01)} & \secondr{0.78 (0.87)} & \secondr{0.21 (0.10)} \\
VN & \textbf{TrackNeRF (ours)} &    & \secondr{18.78} \firstr{(21.45)} & \firstr{0.79 (0.88)} & \firstr{0.20 (0.10)} \\ \bottomrule
\end{tabular}%
}

\label{tab:dtu_gt_pose}
\end{table}

\minorsection{DTU 3 views with ground-truth (GT) poses} 
Many approaches~\cite{freenerf,corresnerf,reconfusion} assume precise GT camera poses available as the sparse view setup, although popular SfM methods like COLMAP~\cite{colmap_sfm} often fail in sparse-view scenario~\cite{sparf2023,dust3r}.
But we also compare our method with representative approaches using different backbone architecture and training settings, to show the effectiveness of our method even using GT poses.
To realize this setup, we fix camera poses to be GT ones and train the NeRF with Eq.~\ref{eq:loss_all}. 
As shown in Tab.~\ref{tab:dtu_gt_pose}, despite vanilla NeRF being used, our method achieves the best performance in PSNR and SSIM, compared with methods using vanilla NeRF or more advanced MipNeRF.
It is worth noting that our method also outperforms diffusion methods like ReconFusion~\cite{reconfusion} and ZeroNVS \cite{zeronvs} trained on large-scale scene data. These results show TrackNeRF improves sparse view rendering quality besides camera pose optimization.

\begin{table}[!htb]
\centering
\caption{\textbf{LLFF Evaluation with No Poses (3 Views)} We show the evaluation on the forward-facing dataset LLFF~\cite{llff} (3 views) with initial identity poses.
Due to the simple camera motion in this dataset, the usage of feature tracks in our method does not lead to significant improvements over SPARF here. 
}
{%
\begin{tabular}{l|cc|ccc}
\toprule
\textbf{Method} & \textbf{Rot. $\downarrow$} & \textbf{Trans. $\downarrow $} & \textbf{PSNR $\uparrow$} & \textbf{SSIM $\uparrow$} & \textbf{LPIPS $\downarrow$} \\

 \midrule
BARF~\cite{barf} & 2.04 & 11.6 & 17.47 & 0.48 & 0.37 \\ 
RegBARF~\cite{Regnerf, barf}  & 1.52 & 5.0 & 18.57 & 0.52 & 0.36 \\
DistBARF~\cite{barf, MipNeRF-360} & 5.59 & 26.5 & 14.69 & 0.34 & 0.49 \\ 
SCNeRF~\cite{SCNeRF} & 1.93 & 11.4 & 17.10 & 0.45 & 0.40 \\
SPARF \cite{sparf2023} & \firstr{0.53} & \secondr{2.8} & \secondr{19.58} & \firstr{0.61} & \firstr{0.31} \\ 
\textbf{TrackNeRF (Ours)} & \secondr{0.77} & \firstr{2.8} &\firstr{19.60} &\secondr{0.59} & \secondr{0.35}  \\ 

\bottomrule
\end{tabular}%
}

\label{tab:llff-pose}
\end{table}

\minorsection{LLFF 3 views without pose} 
For the forward-facing dataset LLFF, we start optimization by identical camera poses. As we can see in Tab.~\ref{tab:llff-pose}, the improvement is not that many compared to DTU, with a slightly better performance on PSNR over SPARF~\cite{sparf2023}.
We think that the reason is that LLFF is a simple dataset without large camera translation and rotation, and thereby the improvement from our feature track consistency and camera pose accuracy are somehow saturated,
as previous works~\cite{nerfminusminus,barf} have shown pose recovery can be done solely from the photometric loss.
Therefore, we conduct our ablation studies mainly on DTU dataset in the following.

\minorsection{Visualization} We visualize the test view renderings from these two datasets to compare the quality of novel view synthesis in Fig.~\ref{fig:vis_main}.
For the DTU dataset with large camera motions, NeRF~\cite{NeRF,nerfminusminus} completely fails while BARF~\cite{barf} cannot recover reasonable poses.
On the contrary, TrackNeRF can still preserve a sharp and high-quality layout compared to SPARF and GT.
Also, we can find clear geometry structure from the rendered depth maps, which indicates that the introduced track optimization effectively helps the radiance field to learn multiview geometry consistency.
Moreover, the floaters and artifacts near the camera or background are also significantly reduced by the introduced depth smoothness loss in Eq.~\ref{eq:depth_reg}.
For the forward-facing LLFF dataset, NeRF~\cite{NeRF,nerfminusminus} is still struggling, while BARF is unable to generate meaningful results.
In contrast, TrackNeRF clearly outperforms these methods in generating views with sharp details, especially in the regions we have highlighted.

\begin{figure}[t]
    \centering
    \begin{subfigure}[b]{\textwidth}
        \includegraphics[width=\linewidth]{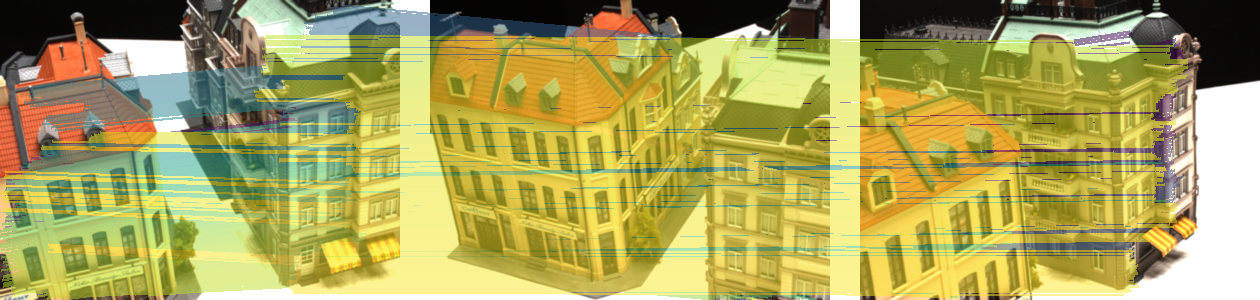}
        \caption{DTU scan 21. An example of a successful case for the correspondence.}  %
        \label{fig:track_vis_a}
    \end{subfigure}

    \vspace{0.2cm}  %

    \begin{subfigure}[b]{\textwidth}
        \includegraphics[width=\linewidth]{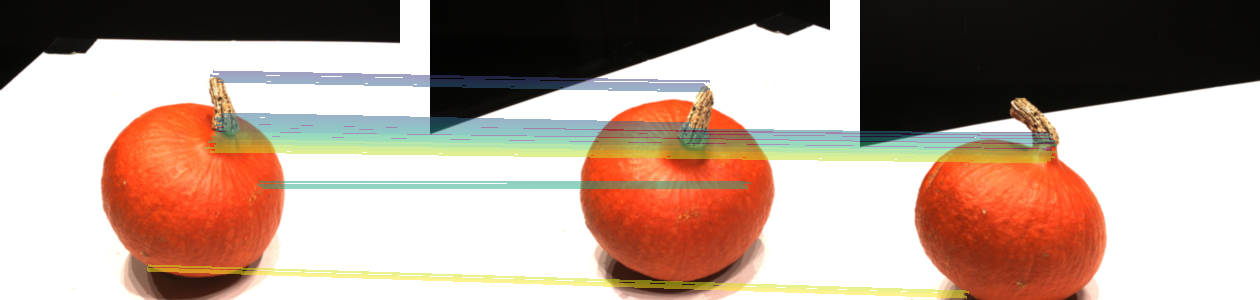}
        \caption{DTU scan 30, The only failed case for the correspondence we find.}  %
        \label{fig:track_vis_b}
    \end{subfigure}
    \caption{ \textbf{Visualization of Feature Tracks.} In almost all scenes, we can always find dense enough and accurate correspondence like the one example from DTU scan 21 shown in Fig.~\ref{fig:track_vis_a}. We provide a rare case (scan 30) in Fig.~\ref{fig:track_vis_b} where the correspondence network~\cite{pdcnet++} fails to find enough reliable correspondences, in which cases we uses a lower $\lambda_{Track}$ for better performance. 
    }
    \label{fig:track_vis}
\end{figure}

\section{Ablation Study and Analysis}\label{sec:ablation}

\begin{table}[!t]
\centering
\caption{
\textbf{Ablation Study on Proposed Components.} 
We conduct our ablation study on DTU and LLFF datasets for all scenes.  
\textbf{\color{black}I}: Coarse to fine frequency encoding introduced by BARF~\cite{barf}; 
\textbf{\color{blue}II}: Track reprojection loss proposed by us in Eq.~\ref{eq:track_loss}; 
\textbf{\color{magenta}III}: Track keypoints refinement in Eq.~\ref{eq:tka}; 
\textbf{\color{olive}IV}: Depth regularization loss in Eq.~\ref{eq:depth_reg}.
}
\resizebox{\textwidth}{!}{%
\begin{tabular}{c|c|cccc|cc|cccc}
\toprule
 & \multirow{2}{*}{\textbf{Method}}
 & \multirow{2}{*}{\color{black}\textbf{I}} 
 & \multirow{2}{*}{\color{blue}\textbf{II}} 
 & \multirow{2}{*}{\color{magenta}\textbf{III}} 
 & \multirow{2}{*}{\color{olive}\textbf{IV}}
 & \multirow{2}{*}{\bf Rot. $\downarrow$} 
 & \multirow{2}{*}{\bf Trans. $\downarrow$}
 & \bf PSNR $\uparrow$ 
 & \bf SSIM $\uparrow$ 
 & \bf LPIPS $\downarrow$ 
 & \bf DE $\downarrow$ \\ 
 &  &  &  &  &  &  &  &  \bf(masked)  &  \bf(masked) & \bf  (masked) &  \\ 
\midrule
\multirow{4}{*}{\rotatebox[origin=c]{90}{DTU}} & BARF & \ding{51} &  &  &   & 10.33 & 51.5 & 10.70 (9.8)    & 0.43 (0.62)& 0.59 (0.36) & 1.9 \\
 & Ours & \ding{51} & {\color{blue}\ding{51}} &  &                       &  1.40 & 2.61 & 18.07 (19.19)  & 0.72 (0.81)& 0.27 (0.14) & 0.11 \\
 & Ours & \ding{51} & {\color{blue}\ding{51}}& {\color{magenta}\ding{51}} &                                   &  1.12 & 2.48 & 18.39 (19.50)  & 0.73 (0.83)& 0.25 (0.13) & 0.11 \\
 & Ours & \ding{51} & {\color{blue}\ding{51}} & {\color{magenta}\ding{51}} & {\color{olive}\ding{51}}                         &  1.12 & 2.48 & 18.53 (19.65)  & 0.73 (0.83)& 0.25 (0.13) & 0.11 \\ \midrule
\multirow{4}{*}{\rotatebox[origin=c]{90}{LLFF}} & BARF & \ding{51} &  &  &  &  2.04 & 11.6 & 17.47          & 0.48       & 0.37         & - \\
 & Ours & \ding{51} & {\color{blue}\ding{51}} &  &                       &  0.80 & 2.96 & 19.34          & 0.58       & 0.36         & - \\
 & Ours & \ding{51} & {\color{blue}\ding{51}} & {\color{magenta}\ding{51}} &                                   &  0.77 & 2.80 & 19.51          & 0.58       & 0.35         & - \\
 & Ours & \ding{51} & {\color{blue}\ding{51}} & {\color{magenta}\ding{51}} & {\color{olive}\ding{51}}                         &  0.77 & 2.80 & 19.60          & 0.59       & 0.35         & - \\ 
 \bottomrule
\end{tabular}%
}
\label{tab:ablations}
\end{table}

\subsection{Ablation on components} 
Starting from BARF~\cite{barf}'s coarse-2-fine encoding inherited by other methods~\cite{freenerf,sparf2023}, we ablate the effectiveness of the key components of our method in Tab.~\ref{tab:ablations}.
The proposed feature track reprojection loss increases the performance significantly, especially on DTU dataset, where the camera motion is large and sparse views are far away from each other with wide baselines.
We achieve $+8$ PSNR boost and also recover much more precise camera poses on DTU.
Since LLFF is a simple forward-facing scene without much camera motion, our improvement is not that much, which aligns with our visual comparison in Fig.~\ref{fig:vis_main}.
Notably, track keypoint adjustment and depth regularization introduced by us also improve the performance further.

\subsection{Ablation on robustness and effectiveness}
\minorsection{Advantage on tolerance of noise level} We follow SPARF's~\cite{sparf2023} settings to certify the robustness of TrackNeRF on pose noise for fair comparison. 
As shown in Tab.~\ref{tab:noise_ablate}, impressively, TrackNeRF is able to converge under $35\%$ of Gaussian noise.
While SPARF\cite{sparf2023} reports failure on $20\%$ noise, TrackNeRF shows stronger robustness to pose noise.
As standard benchmarks always adopt $15\%$ noise as the default scenario, our results show that exploring more challenging pose noise with wider baselines could be more interesting in future directions.

\begin{table}[!htb]
\centering
\caption{ \textbf{Ablation study for noise level on DTU.} We show an ablation study on the noise level for novel view synthesis and camera pose estimation with 3 views in DTU \cite{dtu}. The initial rotation and translation error is obtained by multiplying the noise level with random samples from $\mathcal{N}(\mathbf{0}, \mathbf{I}_6)$ on $\mathfrak{se}(3)$, and then transferring it back to $\text{SE}(3)$.  }
{\begin{tabular}{c|cc|cccc}
\toprule
\textbf{Noise} & \textbf{Rot. $\downarrow$} & \textbf{Trans.$\downarrow$} & \textbf{PSNR $\uparrow$} & \textbf{SSIM $\uparrow$} & \textbf{LPIPS $\downarrow$} & \textbf{DE$\downarrow$} \\ \midrule
0.05 &  0.24 &   0.61 & 15.83 & 0.67 & 0.21 & 0.06 \\
0.15 &  0.22 &   0.60 & 16.14 & 0.67 & 0.21 & 0.07 \\
0.25 &  0.31 &   0.59 & 15.32 & 0.62 & 0.27 & 0.06 \\
0.35 &  0.51 &   1.59 & 14.88 & 0.60 & 0.28 & 0.06 \\
0.45 & 28.28 & 145.07 &  8.34 & 0.32 & 0.54 & 1.84 \\ \bottomrule
\end{tabular}%
}

\label{tab:noise_ablate}
\end{table}

\begin{figure}[!htbp]
    \centering
    \begin{subfigure}[b]{\textwidth}
        \includegraphics[width=\linewidth]{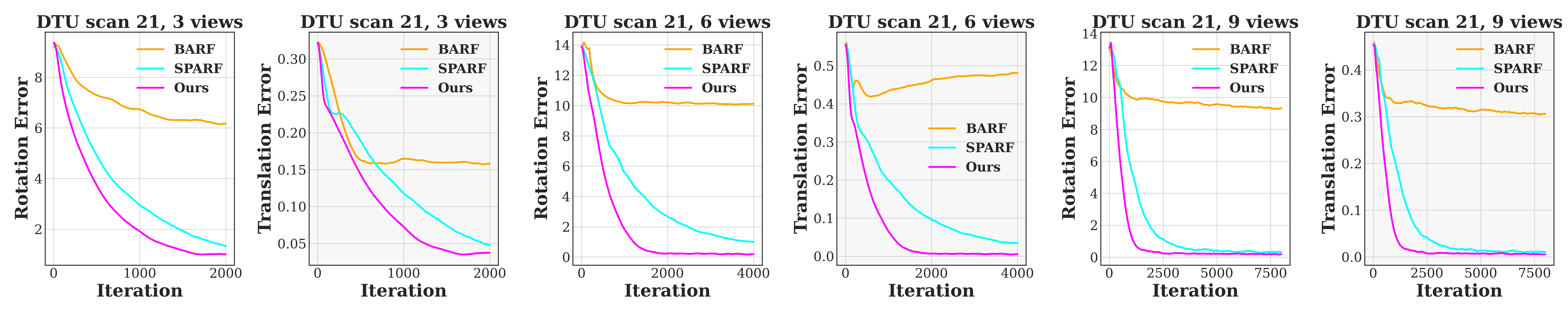}
        \caption{DTU, scan 21}  %
        \label{fig:iter_a}
    \end{subfigure}

    \vspace{0.5cm}  %

    \begin{subfigure}[b]{\textwidth}
        \includegraphics[width=\linewidth]{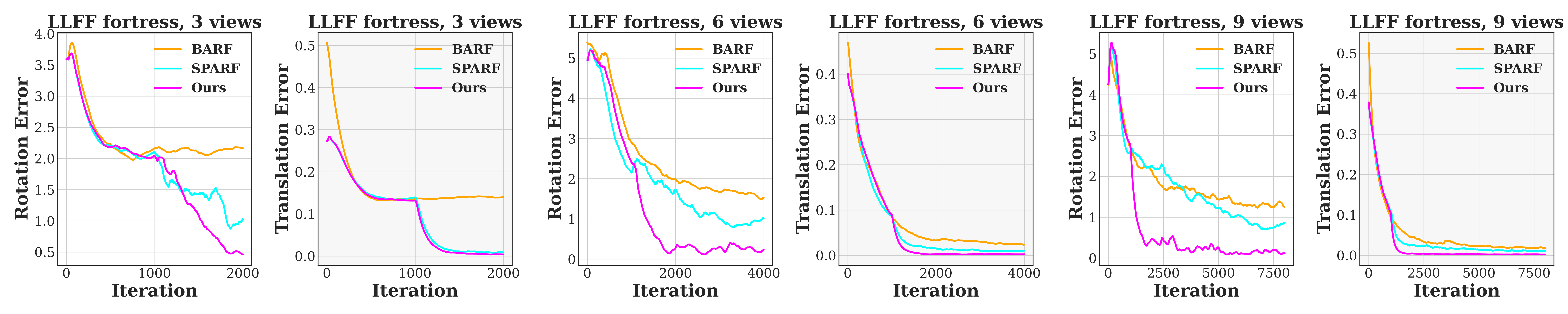}
        \caption{LLFF, fortress scene}  %
        \label{fig:iter_b}
    \end{subfigure}
    \caption{ \textbf{Comparison on the Convergence of Pose Optimization.} We show convergence plots of BARF \cite{barf}, SPARF \cite{sparf2023} and our TrackNeRF on DTU and LLFF datasets.  For a fair comparison, we keep sampling the same number of rays for each iteration as SPARF~\cite{sparf2023}. Plots with white background and \textcolor{gray}{gray} background represent rotation and translation errors, respectively. Our TrackNeRF converges faster and to a lower loss than the state-of-the-art.
    }
    \label{fig:iter}
\end{figure}

\minorsection{Advantage on convergence speed} As shown in Fig.~\ref{fig:iter}, TrackNeRF (\textcolor{deeppurple}{purple curve}) clearly converge faster than BARF and SPARF,
especially under 6-view and 9-view settings, where we have longer feature tracks to guide the pose and underly geometry optimization.
Note for a fair comparison, we force all three methods to sample the same number of rays during each batch.

\subsection{Analysis of feature tracks adequacy}
Due to the sparse view settings for NeRF optimization, concerns may arise regarding the adequacy of feature tracks, both in terms of track number and track length. 
To address these potential concerns, we provide the track statistics as depicted in Fig.~\ref{fig:track_vis}.
Considering the most extreme DTU 3-view case, without generality, NeRF typically samples 2,048 rays per batch for an image with a typical resolution of $(300,400)$. We find an average of $\sim$ 20k tracks of length 3.
Given these statistics, it is evident that the density of feature tracks is sufficient to facilitate robust NeRF optimization without the risk of excessive sampling redundancy. This surfeit of feature tracks ensures that the optimization process is well-supported by adequate data, even in the most extreme sparse view conditions.

\section{Limitations}
A noteworthy scenario arises when only two views are available.
Under these rather rare circumstances, our track loss reverts to an optimization based on pairwise correspondences.
However, we demonstrate the superiority of our proposed track optimization approach, particularly within the context of a \textit{moderate sparse} setting, such as the case with six noisy views in Tab.~\ref{tab:dtu_noise_pose}.
Another limitation occurs when the matching network fails to get correct and sufficient correspondence as shown in Fig.~\ref{fig:track_vis_b}, which is also shared by ~\cite{sparf2023,corresnerf}. We believe this can be addressed by leveraging more advanced feature matchers, which we leave for future exploration.

\section{Conclusions and Future Work}
We propose TrackNeRF for novel view synthesis under sparse and noisy views.
Our method introduces feature track optimization for joint learning of camera poses and neural radiance field.
Our joint optmization seamlessly aligns with the objective of bundle adjustment (BA) that has been widely adopted in SfM as a golden practice for decades.
TrackNeRF can tolerate greater pose noise and converge faster, benefiting from global BA.
We also show that TrackNeRF can achieve higher rendering quality, restore more accurate poses, and even outperform advanced diffusion-based methods trained on large-scale datasets.
Overall, we show that multiview correspondence is critical in sparse view settings, especially with inaccurate or without poses.
Moreover, our contribution is orthogonal to those learning-based methods and can be easily integrated into them without bells and whistles. 
We believe TrackNeRF can inspire future interesting research and have further impacts on the community.

\minorsection{Acknowledgement}
The research reported in this publication was supported by funding from KAUST Center of Excellence on GenAI under award number 5940,
as well as, the SDAIA-KAUST Center of Excellence in Data Science and Artificial Intelligence (SDAIA-KAUST AI). Part of the support is also coming from the KAUST Ibn Rushd Postdoc Fellowship program.

\clearpage

\renewcommand{\thesection}{\Alph{section}}
\renewcommand{\thetable}{\Roman{table}}
\renewcommand{\thefigure}{\Roman{figure}}
\setcounter{section}{0}
\setcounter{table}{0}
\setcounter{figure}{0}

\renewcommand{\theequation}{\thesection.\arabic{equation}}
\setcounter{equation}{0}

\begin{center}
{\Large\bf TrackNeRF: Bundle Adjusting NeRF from Sparse and Noisy Views via Feature Tracks
}\\
\vspace{1em}
{\large\bf--- Supplementary Material ---}
\end{center}

In this appendix, we provide additional content to complement the main manuscript:
\begin{itemize}[leftmargin=1em,topsep=0pt]

\item \applabel~\ref{apx:add_exp}: Additional results and experiments, including different experiment settings, components ablation, and detailed per-scene results.
\item \applabel~\ref{apx:add_discuss}: Detailed discussion and comparisons to existing literature that we can't put in the main paper due to limited space.
\end{itemize}

\section{Additional Experiments}~\label{apx:add_exp}

We present additional results and experiments, including different experiment settings, components ablation, runtime analysis, detailed per-scene results, and more fine-grained visualizations.

\subsection{Per-scene Results}

For loss weights, we usually fix $\lambda_{Depth}=10^{-1},\lambda_{Track}=10^{-3}$. But we gradually anneal $\lambda_{Track}$ to $10^{-5}$ when matcher failed, since we usually find $>10k$ tracks but only $\sim0.1k$ unconfident tracks in such a case. 
We also provide the per-scene results in Tab.~\ref{tab:per_scene_dtu}.

\begin{table}[!htb]
\centering
\resizebox{\textwidth}{!}{%
\begin{tabular}{@{}cccccccccc@{}}
\toprule
\textbf{Scene} & \textbf{Rot. (degree)$\downarrow$} & \textbf{Trans. (x100)$\downarrow$} & \textbf{PSNR $\uparrow$} & \textbf{(masked)} & \textbf{SSIM $\uparrow$} & \textbf{(masked)} & \textbf{LPIPS $\downarrow$} & \textbf{(masked)} & \textbf{DE$\downarrow$} \\ \midrule
scan8 & 0.119 & 0.350 & 18.541 & 20.256 & 0.762 & 0.837 & 0.251 & 0.167 & 0.040 \\
scan21 & 0.222 & 0.603 & 15.564 & 16.135 & 0.602 & 0.666 & 0.264 & 0.211 & 0.073 \\
scan30 & 12.017 & 26.499 & 12.449 & 11.969 & 0.678 & 0.880 & 0.442 & 0.153 & 0.441 \\
scan31 & 0.438 & 1.090 & 16.640 & 18.201 & 0.706 & 0.777 & 0.197 & 0.127 & 0.060 \\
scan34 & 0.306 & 1.000 & 18.147 & 20.728 & 0.698 & 0.753 & 0.248 & 0.195 & 0.043 \\
scan38 & 0.203 & 0.709 & 16.604 & 18.758 & 0.637 & 0.707 & 0.290 & 0.235 & 0.053 \\
scan40 & 0.076 & 0.191 & 15.260 & 18.037 & 0.597 & 0.709 & 0.331 & 0.209 & 0.108 \\
scan41 & 0.102 & 0.226 & 15.727 & 19.941 & 0.693 & 0.787 & 0.279 & 0.168 & 0.069 \\
scan45 & 0.443 & 0.764 & 15.323 & 17.716 & 0.731 & 0.816 & 0.197 & 0.128 & 0.089 \\
scan55 & 0.191 & 0.545 & 22.471 & 22.202 & 0.749 & 0.881 & 0.236 & 0.066 & 0.096 \\
scan63 & 0.673 & 1.476 & 24.322 & 22.703 & 0.892 & 0.948 & 0.124 & 0.037 & 0.110 \\
scan82 & 0.340 & 0.711 & 19.709 & 20.401 & 0.861 & 0.934 & 0.162 & 0.053 & 0.163 \\
scan103 & 0.134 & 0.564 & 21.189 & 22.611 & 0.807 & 0.910 & 0.209 & 0.085 & 0.083 \\
scan110 & 1.344 & 1.989 & 21.694 & 20.671 & 0.760 & 0.898 & 0.273 & 0.072 & 0.148 \\
scan114 & 0.162 & 0.513 & 24.254 & 24.452 & 0.845 & 0.927 & 0.198 & 0.066 & 0.024 \\ \bottomrule
\end{tabular}%
}
\caption{\textbf{DTU Evaluation (3 Noisy Views)}. We benchmark novel view synthesis and camera pose estimation methods on DTU \cite{dtu} with noisy initial poses with 3 input views. We simulate noisy poses by adding $15\%$ of Gaussian noise to the ground-truth poses. 
Rotation errors are in degree and translation errors are multiplied by 100. Results in ($\cdot$) are computed by masking the background.}
\label{tab:per_scene_dtu}
\end{table}

\subsection{Runtime efficiency}
We ablate and discuss the runtime efficiency of our method in this section.
In fact, the matching and track extraction is fast and can be done within seconds. The main time cost comes from vanilla NeRF~\cite{NeRF} optimization, as we're inheriting vanilla slow NeRF/BARF for fair comparisons. 
Since we use different MLP backbones as SPARF on LLFF(coarse-only) and DTU(coarse-fine), we provide the detailed time and performance comparisons in Tab.~\ref{tab:depth_stage} for DTU dataset.
Theoretically, because we always render $N$ pixels for each iteration, suppose the average feature track length is $\bar{l}$, then the reprojection loss has

\begin{equation}
    \text{num\_track}\times \text{proj\_per\_track}=\frac{N}{\bar{l}}\times 2C_{\bar{l}}^{2}=N(\bar{l}-1)=\mathcal{O}(N\bar{l})
\end{equation}

terms to back-propagate. SPARF has a fixed $\bar{l}=2$ since it samples paired images, but we'll have $2\le \bar{l} \le n$ since we consider holistic feature tracks from $n$ views, which results in more time per iteration.
The reported Ours (60K) with the best performance is trading off longer cumulative time to surpass SPARF for all metrics.
However, by reducing training iterations, we highlight that even with 88\% training time of SPARF, Ours(23K) is able to outperform SPARF on both pose accuracy and PSNR.
With efficient rendering methods like Gaussian splatting~\cite{gaussiansplatter} or InstantNGP~\cite{InstantNGP}, we believe our optimization cost can be dramatically reduced further, which we leave for future exploration.

\begin{table}[t]
\centering

\vspace{-1em}
{%
\begin{tabular}{c|c|cc|ccc}
\toprule
\textbf{Method} & \textbf{Time} & \textbf{Rot.} & \textbf{Trans.} & \textbf{PSNR} & \textbf{SSIM}   & \textbf{DE} \\

\midrule

BARF & 5.6h $(1.0\times)$ & 10.33 & 51.5 & 9.76 & 0.62 & 1.90 \\ 
\cellcolor{gray!10} SPARF & \cellcolor{gray!10} 11.6h $ \cellcolor{gray!10} (2.1\times)$ & \cellcolor{gray!10} 1.81 & \cellcolor{gray!10} 5.0 & \cellcolor{gray!10} 18.92 & \cellcolor{gray!10} \secondr{0.83}  & \cellcolor{gray!10} \secondr{0.12} \\

\cellcolor{gray!25} Ours (23K) & \cellcolor{gray!25} 10.3h $(1.8 \times)$          & \cellcolor{gray!25} \firstr{1.11} & \cellcolor{gray!25} \secondr{3.61} & \cellcolor{gray!25} \secondr{19.35} & \cellcolor{gray!25} 0.81  & \cellcolor{gray!25} \secondr{0.12} \\
\textbf{Ours (60K)}  & 27.4h  $(4.9 \times)$       & \secondr{1.12} & \firstr{2.48} & \firstr{19.65} & \firstr{0.83}  & \firstr{0.11} \\
 \midrule

$\mathcal{L}_{Depth}^{Stage1}$+$\mathcal{L}_{Depth}^{Stage2}$      & 27.6h $(4.9 \times)$      & 1.49 & 3.12 & 19.57 & 0.82 & 0.10 \\

\bottomrule

\end{tabular}%
}
\caption{ \textbf{Mid-rows: Training time.} Ours (60K iterations) is the default TrackNeRF we reported in the paper, which requires a longer time for training. Ours (23K iterations) is a more efficient version of TrackNeRF by reducing optimization iterations. \textbf{Last-row: $\mathcal{L}_{Depth}^{Stage1}$+$\mathcal{L}_{Depth}^{Stage2}$.} Adding depth loss from Stage 1 until the end of Stage 2.   } 
\vspace{-2em}
\label{tab:depth_stage}
\end{table}

\begin{table}[!htb]
\centering
\caption{\textbf{DTU Evaluation with 0.35 Noise (3 Views)} We show the evaluation on the DTU~\cite{dtu} dataset (3 views) with initial poses perturbed by $35\%$ gaussian noise. The results come from all the scenes in DTU.
}
{%
\begin{tabular}{l|cc|cccc}
\toprule
\textbf{Method} & \textbf{Rot. $\downarrow$} & \textbf{Trans. $\downarrow $} & \textbf{PSNR $\uparrow$} & \textbf{SSIM $\uparrow$} & \textbf{LPIPS $\downarrow$}  & \textbf{DE $\downarrow$} \\
 \midrule
BARF~\cite{barf} & 27.85 & 68.48 & 8.72(9.37) & 0.29(0.60) & 0.73(0.38) & 1.10 \\
SPARF \cite{sparf2023} & \secondr{5.26} & \secondr{19.39} & \secondr{15.51(16.12)} & \secondr{0.62(0.75)} & \secondr{0.35(0.21)} & \secondr{0.31} \\ 
\textbf{TrackNeRF (Ours)} & \firstr{3.49} & \firstr{11.12} &\firstr{16.21(16.77)} &\firstr{0.66(0.79)} & \firstr{0.33(0.18)} & \firstr{0.17}  \\ 

\bottomrule
\end{tabular}%
}

\label{tab:noise_35}
\end{table}

\subsection{Robustness to Noisy Camera Pose}
\begin{figure}[!htb]
    \centering
    \includegraphics[width=\linewidth]{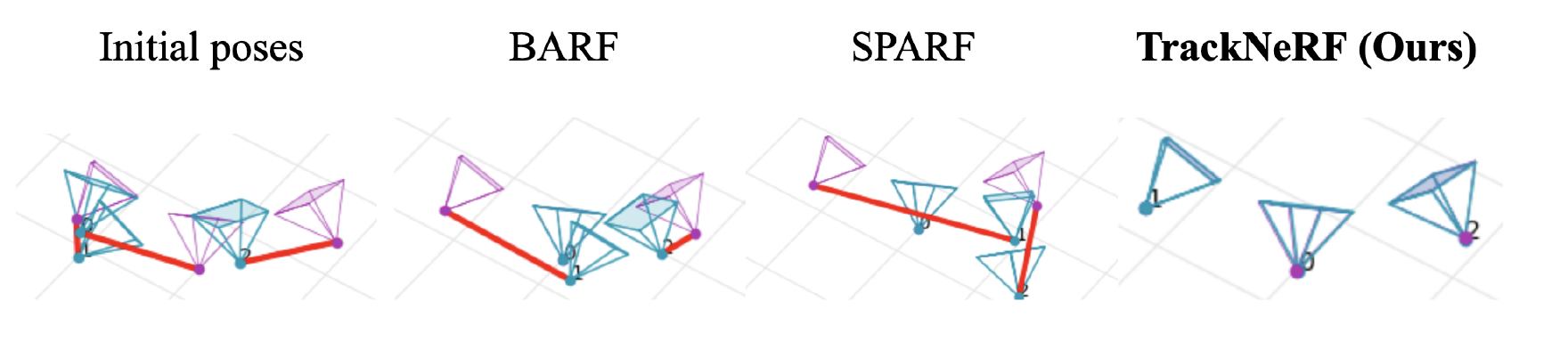}
    \caption{
    \textbf{Visualization of Camera Poses under 35\% of Noise } Only our TrackNeRF can recover near perfect camera poses even under $35\%$ of noise in this selected example.
    }
    \label{fig:vis_pose_035}
\end{figure}

We show the proposed TrackNeRF can tolerate up to 0.35 pose noise on one example scene from robustness experiments in the main paper.
Here, we further challenge our method and evaluate the robustness of our method to noise by increasing the noise to 0.35 for all the scenes in DTU. As shown in Tab.~\ref{tab:noise_35}, thanks to our global correspondence-based TrackNeRF, our method outperforms SPARF and BARF by a larger margin, compared to standard 0.15 noise.
We find the improvements mainly come from some challenging scenes where SPARF stucks at sub-optimal while TrackNeRF converges successfully. 
Fig.~\ref{fig:vis_pose_035} illustrates the visualized example of camera pose. Please refer to Sec.~\ref{sec:ba_discussion} for the explanation of why our global consistency can converge in more challenging settings

\begin{figure}[!htb]
    \centering
    \includegraphics[width=\linewidth]{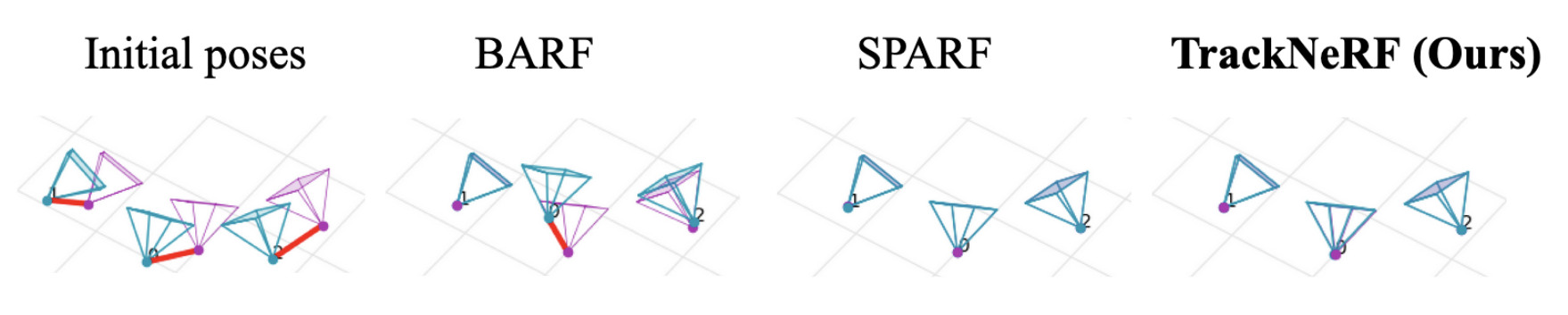}
    \caption{
    \textbf{Visualization of Camera Poses under 15\% of Noise,} where {\color{teal} teal} and {{\color{Purple} purple}} tetrahedrons indicate recovered and ground-truth camera pose, respectively.  While BARF~\cite{barf} can't recover the camera poses well, both SPARF~\cite{sparf2023} and our TrackNeRF can recover near perfect camera poses under $15\%$ of noise.
    }
    \label{fig:vis_pose_015}
\end{figure}

\subsection{Visualization of Robustness Experiments}
\begin{figure}[!htb]
    \centering
    \includegraphics[width=\linewidth]{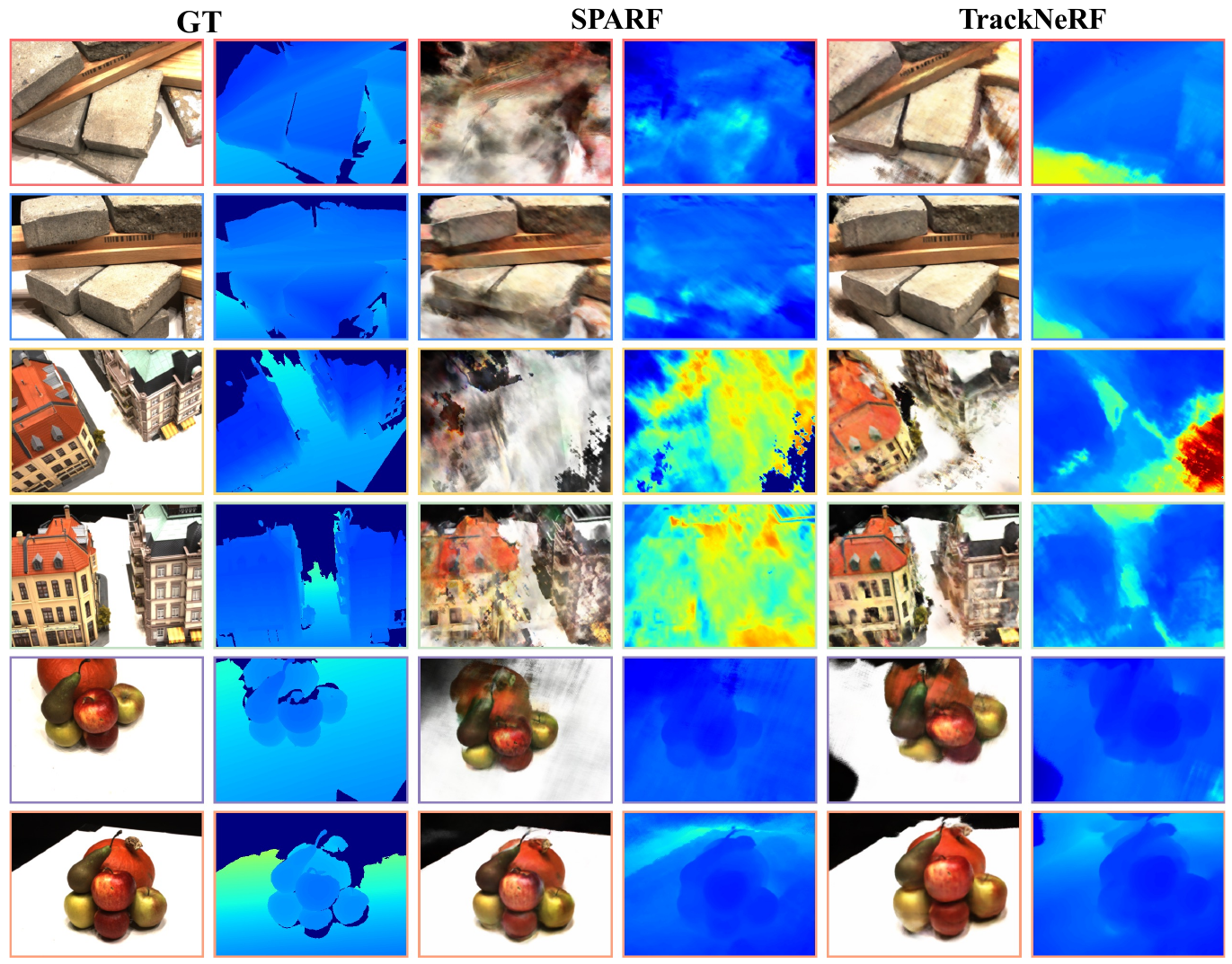}
    \caption{
    \textbf{Visualized comparison under noise=0.35 on DTU (3 Views).} We don't cherrypick specific views here. We select $1^{th}$ and $24^{th}$ view from the test view split since it has 48 views in total.
    }
    \label{fig:vis_main_35}
\end{figure}

Fig.~\ref{fig:vis_pose_015} and Fig.~\ref{fig:vis_pose_035}  visualize the camera poses optimized by BARF, SPARF and our method. As shown in Fig.~\ref{fig:vis_pose_015},  both SPARF~\cite{sparf2023} and our TrackNeRF can recover near-perfect camera poses for camera pose with $15\%$ noise. However, when pose noise is increased to $35\%$, our method achieves the best performance in recovering camera pose,  compared with SPARF and BARF, as shown in Fig.~\ref{fig:vis_pose_035}. TrackNeRF also achieves better visual quality, as shown in Fig.~\ref{fig:vis_main_35}.

\subsection{Effect of Track Length}

\begin{figure}[!htbp]
    \centering
    \begin{subfigure}[b]{\textwidth}
        \includegraphics[width=\textwidth]{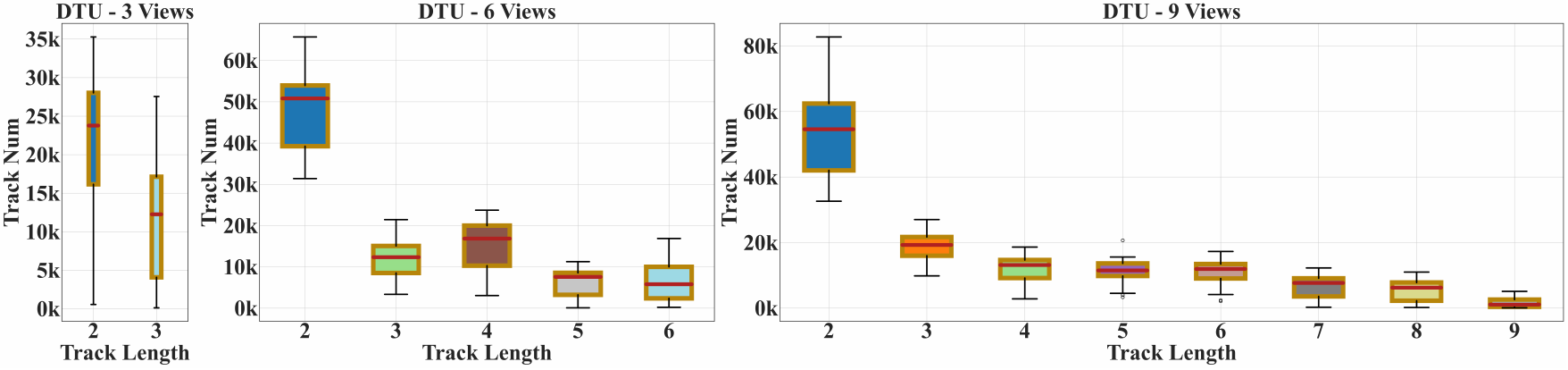}
        \caption{DTU number of tracks}  %
        \label{fig:track_num_a}
    \end{subfigure}

    \vspace{0.5cm}  %

    \begin{subfigure}[b]{\textwidth}
        \includegraphics[width=\textwidth]{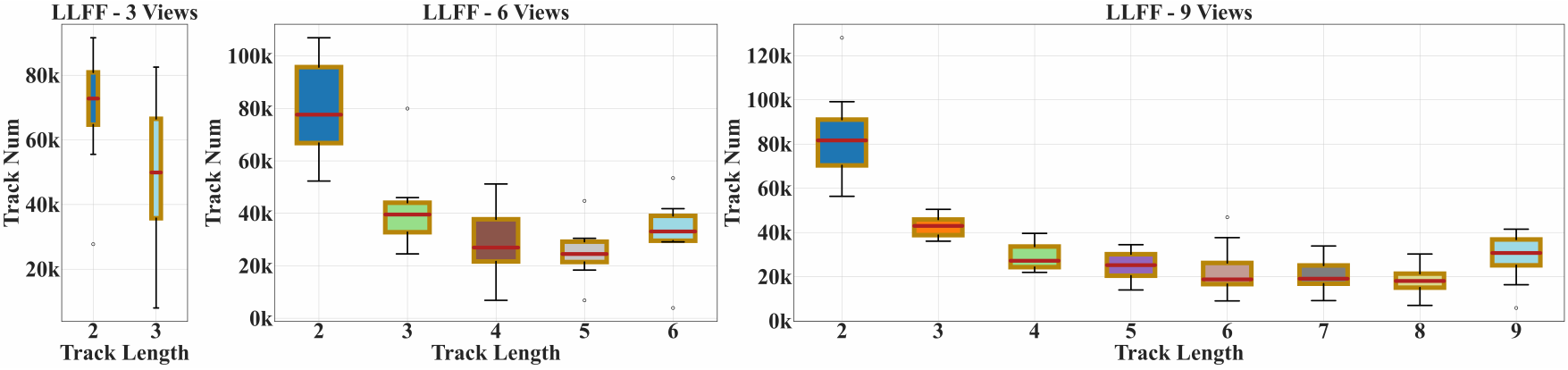}
        \caption{LLFF number of tracks}  %
        \label{fig:track_num_b}
    \end{subfigure}
    \caption{\textbf{Distribution of Tracks' Length in TrackNeRF}. We show the frequency of each track's lengths of TrackNeRF in the DTU~\cite{dtu} and LLFF~\cite{llff} datasets under 3, 6, and 9 views. Note that even for sparse views as less as 3 views, it's still possible to find enough tracks for our track-based optimization.
    }
    \label{fig:track_num}
\end{figure}

\minorsection{Track Length statistics} To address the concern that there are not enough long tracks for our proposed feature-track-based optimization,
we compute the length of feature tracks for both the DTU and LLFF datasets. Fig.~\ref{fig:track_num} shows there are always enough tracks for our feature track optimization.

\minorsection{Effects of Track Length} SPARF could be a particular case of TrackNeRF where only tracks with a length of two are constructed and optimized. Thus, to show the effect of track lengths on the optimization, we progressively add the tracks with different lengths in the DTU-6 view setup.
As listed in Tab.~\ref{tab:track_len}, adding longer feature tracks into the optimization leads to better  PSNR value and more accurate camera pose.

\begin{table}[!htb]
\centering
\caption{ \textbf{Ablation study for Track Length on DTU.} We show an ablation study on the track length for novel view synthesis and camera pose estimation with 6 views in DTU \cite{dtu}. We randomly pick a scene from DTU, and restrict the max length of the feature tracks to $2,3,4,5,6$, respectively. As longer feature tracks are allowed to join the optimization, the performance for both view synthesis and pose registration is improving.}
{\begin{tabular}{c|cc|cccc}
\toprule
\textbf{Track Len. $\le$} & \textbf{Rot. $\downarrow$} & \textbf{Trans.$\downarrow$} & \textbf{PSNR $\uparrow$} & \textbf{SSIM $\uparrow$} & \textbf{LPIPS $\downarrow$} & \textbf{DE$\downarrow$} \\ \midrule
2 &  5.92 &   23.01 & 13.15(13.11) & 0.42(0.50) & 0.45(0.38) & 0.27 \\
3 &  0.10 &   0.32 & 19.10(19.14) & 0.73(0.76) & 0.19(0.15) & 0.03 \\
4 &  0.16 &   0.39 & 19.20(19.24) & 0.73(0.77) & 0.19(0.15) & 0.03 \\
5 & 0.14 & 0.40 & 19.10(19.29) & 0.74(0.78) & 0.17(0.14) & 0.03 \\
6 &  0.09 &   0.36 & 19.56(19.46) & 0.74(0.78) & 0.18(0.15) & 0.03 
\\ \bottomrule
\end{tabular}%
}

\label{tab:track_len}
\end{table}

\subsection{Track Keypoint Adjustment}

To validate the effectiveness of track keypoint adjustment (TKA), we also incorporate it with a baseline method SPARF.
To implement this, we extract all the feature tracks, and then perform TKA to optimize the feature matches on a track-wise objective.
After that, we decompose all these feature tracks with various lengths into fixed 2-view correspondences, on which SPARF's optimization works.
As shown in Tab.~\ref{tab:sparf_tka}, our TKA improves SPARF in multiple metrics \eg PSNR, SSIM and LPIPS.

\begin{table}[!htb]
\centering
\caption{\textbf{Track Keypoint Optimization (TKA) for SPARF} 
}
{%
\begin{tabular}{l|cc|cccc}
\toprule
\textbf{Method} & \textbf{Rot. $\downarrow$} & \textbf{Trans. $\downarrow $} & \textbf{PSNR $\uparrow$} & \textbf{SSIM $\uparrow$} & \textbf{LPIPS $\downarrow$}  & \textbf{DE $\downarrow$} \\
 \midrule
 SPARF \cite{sparf2023} & 1.81 & 5.0 & 17.74 (18.92) & 0.71 \secondr{(0.83)} & 0.26 (0.13)& \secondr{0.12} \\
 
SPARF+TKA  & \secondr{1.38} & \secondr{3.99} & \secondr{17.84(19.15)} & \secondr{0.72}(0.82) & \secondr{0.25}\firstr{(0.12)} & \secondr{0.12} \\ 

\textbf{TrackNeRF (Ours)}  & \firstr{1.12} & \firstr{2.48} & \firstr{18.53 (19.65}) & \firstr{0.73 (0.83)} & \firstr{0.25}\secondr{ (0.13)} & \firstr{0.11} \\

\bottomrule
\end{tabular}%
}

\label{tab:sparf_tka}
\end{table}

\begin{figure}[!htb]
    \centering
    \includegraphics[width=\linewidth]{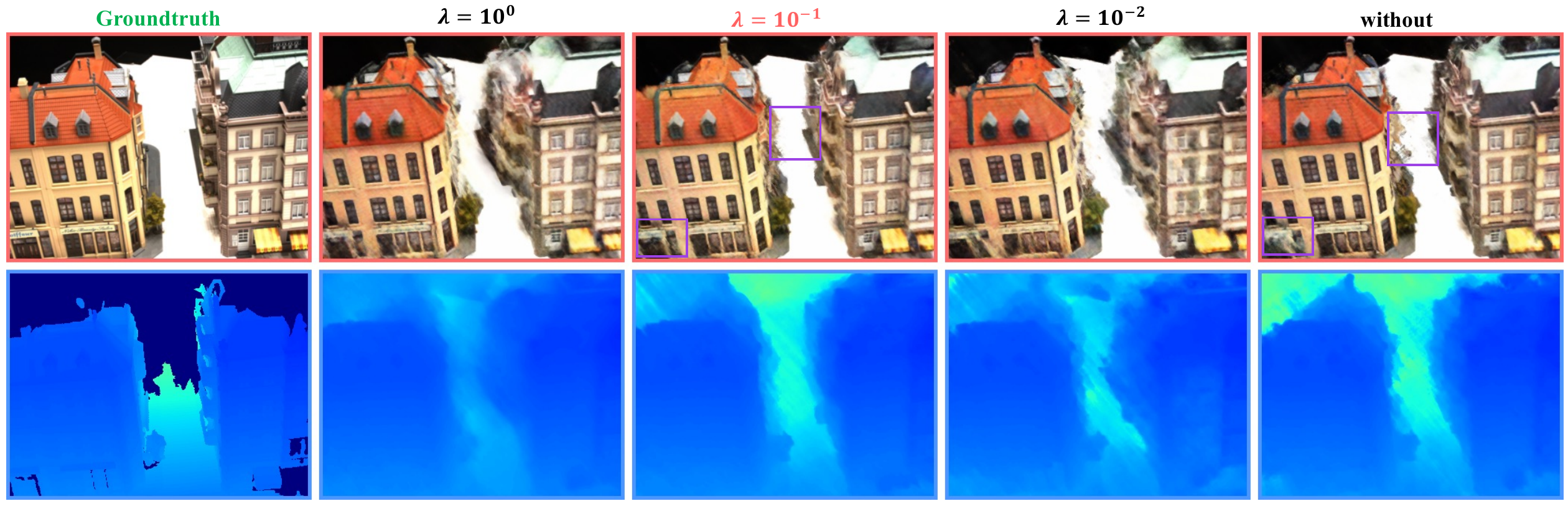}
    \caption{
    \textbf{Effect of Depth Regularization Loss.} The setting of $\lambda_{depth}=10^{-1}$ achieves the best result in terms of both visual and numeric quality.
    }
    \label{fig:depth_reg}
\end{figure}

\subsection{Effect of depth regularization}

We exponentially vary the weight $\lambda_{depth}$ of depth regularization loss by ten. As shown in Fig.~\ref{fig:depth_reg}, $\lambda_{depth}=10^{0}$ results in an over-smoothed result, making it difficult to distinguish the foreground buildings from the background in the depth map. In contrast, $\lambda_{depth}=10^{-2}$ improperly encourages the depth map to be over-coarse. We empirically find the setting of $\lambda_{depth}=10^{-1}$ achieves the highest image quality of rendered views, especially compared with the result without any depth regularization (see marked regions in Fig.~\ref{fig:depth_reg}). These results show that the depth regularization loss can effectively reduce the ghost-like artifacts and floaters.
Tab. \ref{tab:depth_stage} shows applying $\mathcal{L}_{Depth}$ for both stages degrades our performance.

\section{More Discussions}~\label{apx:add_discuss}

This section provides more detailed discussions and comparisons with previous works.

\subsection{Comparison with BARF}

BARF~\cite{barf}, entitled with \textit{"Bundle-Adjusting Neural Radiance Fields"}, might be the first paper that proposes to do bundle adjustment (BA) inside the neural radiance field.
They give valuable insight that the naive position encoding strategy proposed by NeRF~\cite{NeRF} may lead to a suboptimal camera registration solution due to the optimization's complex backward signal. Therefore, they propose a simple yet effective coarse-to-fine position encoding strategy to guarantee that smooth signals will be passed during the optimization. They gradually active the encodings to higher frequency until full positional encoding, which has been adopted in the following works, including FreeNeRF~\cite{freenerf}, SPARF~\cite{sparf2023} and our TrackNeRF.

However, BARF's main contribution is to propose coarse-to-fine encoding and perform a similar task to register poses like BA. They don't introduce any reprojection constraint like the formulation of de facto BA. 

\subsection{Comparison with SPARF} 

\minorsection{Local BA and Global BA} \label{sec:ba_discussion}
As discussed in the main paper, SPARF~\cite{sparf2023} iteratively selects two views, then considers all pairwise correspondence and optimizes the two-view reprojection error. In this way, their method shares similarities to \textit{local bundle adjustment}, which can be dated back many works~\cite{real_local_ba} from SLAM/incremental SfM. Local BA usually limits the reprojection error~\cite{local_ba_err} to involve \textbf{(1)} $n$ most recent frames or \textbf{(2)} features appeared in $N$ last frames.

On the contrary,  our method optimizes sampled holistic feature tracks from all the views for global correspondence consistency, sharing more similarities to the standard (global) BA~\cite{global_ba_modern}. Both local BA and global BA can converge to good minimums when the initial cost is small given enough iterations, as the close performance between SPARF and ours in LLFF~\cite{llff} experiments have shown. 
However, our method additionally offers strengths in two aspects:

\begin{enumerate}[label=\textbf{(\arabic*.)}]
    \item  In this global way, we can reduce the uncertainty and error propagations suffered by local BA~\cite{local_ba_err}. Our TrackNeRF can also register more challenging camera poses and avoid suboptimal minimum, as our noise robustness experiments in Tab.~\ref{tab:noise_35} show.
    \item TrackNeRF can explicitly consider underlying consistency. For example, if the correspondence network only finds correspondences $(u,v)$ and $(v,w)$ without $(w,u)$, TrackNeRF can still ensure the consistency for $(w,u)$ explicitly.
\end{enumerate}

Since we are assuming sparse view settings in NeRF training, we argue that restricting to two views is not necessary. We can easily scale up globally with consideration of correspondence to all the feature tracks and corresponding views for the benefits we discussed above.

Note that we use the term \textit{similarity} here when talking about local BA and global BA. The similarity majorly lies in the objective of the loss function. Because both SPARF and ours are sampling \textit{batches} to perform gradient descent with neural radiance field (MLP) representation, where classic BA usually takes all the optimizable parameters into cost function, then performs nonlinear optimization using Gauss-Newton or Levenberg-Marquadt method, with sparse and explicit 3D points as scene representation.

\subsection{Comparison with other NeRFs with BA} 
There are also many other neural radiance field papers entitled with \textit{bundle adjustment}. 
However, as most of them are not closely related to the reprojection loss of BA, we put the comparison and discussion with them here.

Both USB-NeRF~\cite{usb_nerf} and BAD-NeRF~\cite{bad_nerf} focus on utilizing characteristics of physical cameras to optimize NeRF and camera poses.
L2G-NeRF~\cite{l2g_ba_nerf} learns to transform local coordinates to global and then optimizes camera pose by differentiable parameter estimation in the global coordinate system with photometric loss~\cite{photometric_ba_1,photometric_ba_2}.
UC-NeRF~\cite{uc_nerf} performs the spatio-temporal pose refinement  with the photometric loss for dense multi-camera video input in autonomous driving.
GARF~\cite{GARF} proposes an embedding-free NeRF with Gaussian activation.
CBARF~\cite{cbarf} proposes a neighbor-replacement and coarse-to-fine strategy to improve BARF~\cite{barf}.
DBARF~\cite{dbarf} takes the scene graph as input, adopts an implicit objective based on deep feature cost map, and optimizes generalizable NeRF with pose through GRU~\cite{droid_slam,raft,raft3d} net.
SaNeRF~\cite{chen2022structure} exploits the correspondence reprojection loss in 3D space to train a pose network for pose optimization.
CoPoNeRF~\cite{coponerf} unifies correspondence and NeRF to train an end-to-end framework for generalizable pose-free NeRFs.
DReg-NeRF~\cite{dreg_nerf} registers different NeRFs through learning to align 3D feature volumes.
UP-NeRF~\cite{up_nerf} optimizes NeRF and pose through deep feature alignment.
Some works~\cite{pf_lrm,leap} also train an end-to-end model to directly predict NeRF from unposed images.

\subsection{Comparison with other NeRFs with correspondence}
Many Neural radiance field works (\eg \cite{match_nerf,consistent_nerf,cmc_nerf,re_nerfing}) utilize multiview correspondence like CorresNeRF~\cite{corresnerf}. However, since most of them are not doing pose optimization as the purpose of BA, we discuss these methods here.
MatchNeRF~\cite{match_nerf} uses explicit correspondence for novel view synthesis from generalizable NeRFs.
ConsistentNeRF~\cite{consistent_nerf} also utilizes left-view and right-view reprojection consistency, along with additional depth supervision, for sparse-view NeRF.
SfMNeRF~\cite{chen2023improving} optimizes pairwise left-right reprojection loss as well, but they project interpolated patches into 3D as the cost term, along with depth smooth loss and epipolar loss.
CMC~\cite{cmc_nerf} proposes cross-view multiplane feature consistency for few-shot novel view synthesis.
Re-Nerfing~\cite{re_nerfing} proposes to train NeRF again by enforcing the geometric constraints.

\clearpage

\bibliographystyle{splncs04}
\bibliography{main}
\end{document}